\documentclass[sigconf, nonacm]{acmart}

\newcommand\vldbdoi{XX.XX/XXX.XX}
\newcommand\vldbpages{XXX-XXX}
\newcommand\vldbvolume{14}
\newcommand\vldbissue{1}
\newcommand\vldbyear{2020}
\newcommand\vldbauthors{\authors}
\newcommand\vldbtitle{\shorttitle} 

\newcommand\vldbavailabilityurl{https://github.com/CoDS-GCS/kgpip-public}

\newcommand\vldbpagestyle{plain} 

\usepackage{soul}
\usepackage{subcaption}
\usepackage{amsmath}

\usepackage[linesnumbered,ruled]{algorithm2e}

\newcommand{\sysname}{KGpip}
\newcommand{\sysGC}{GraphGen4Code}
\newcommand{\GML}{MetaPip}

\newcommand{\hitext}[1]{\textcolor{black}{#1}}

\def\ncp{\vspace*{-0.5ex}}

\usepackage{booktabs} 
\usepackage{makecell}

\begin{document}
\title{A Scalable AutoML Approach Based on Graph Neural Networks}

\author{Mossad Helali$^\ast$, Essam Mansour$^\ast$, Ibrahim Abdelaziz$^\S$, Julian Dolby$^\S$, Kavitha Srinivas$^\S$}
\affiliation{%
  \institution{ $^\ast$Concordia University \qquad\qquad\qquad $^\S$IBM Research AI}
  \city{Montreal, Canada \qquad\qquad\qquad\qquad New York, US }
 }
 \email{{fname}.{lname}@concordia.ca, ibrahim.abdelaziz1, dolby, Kavitha.Srinivas@ibm.com}







\begin{abstract}
AutoML systems build machine learning models automatically by performing a search over valid data transformations and learners, along with hyper-parameter optimization for each learner. Many AutoML systems use meta-learning to guide search for optimal pipelines.  In this work, we present a novel meta-learning system called KGpip which (1) builds a database of datasets and corresponding pipelines by mining thousands of scripts with program analysis, (2) uses dataset embeddings to find similar datasets in the database based on its content instead of metadata-based features, (3) models AutoML pipeline creation as a graph generation problem, to succinctly characterize the diverse pipelines seen for a single dataset. KGpip's meta-learning is a sub-component for AutoML systems. We demonstrate this by integrating KGpip with two AutoML systems. Our comprehensive evaluation using 121 datasets, including those used by the state-of-the-art systems, shows that KGpip significantly outperforms these systems.
\end{abstract}


\maketitle

\pagestyle{\vldbpagestyle}
\begingroup\small\noindent\raggedright\textbf{PVLDB Reference Format:}\\
\vldbauthors. \vldbtitle. PVLDB, \vldbvolume(\vldbissue): \vldbpages, \vldbyear.\\
\href{https://doi.org/\vldbdoi}{doi:\vldbdoi}
\endgroup
\begingroup
\renewcommand\thefootnote{}\footnote{\noindent
This work is licensed under the Creative Commons BY-NC-ND 4.0 International License. Visit \url{https://creativecommons.org/licenses/by-nc-nd/4.0/} to view a copy of this license. For any use beyond those covered by this license, obtain permission by emailing \href{mailto:info@vldb.org}{info@vldb.org}. Copyright is held by the owner/author(s). Publication rights licensed to the VLDB Endowment. \\
\raggedright Proceedings of the VLDB Endowment, Vol. \vldbvolume, No. \vldbissue\ %
ISSN 2150-8097. \\
\href{https://doi.org/\vldbdoi}{doi:\vldbdoi} \\
}\addtocounter{footnote}{-1}\endgroup

\ifdefempty{\vldbavailabilityurl}{}{
\vspace{.3cm}
\begingroup\small\noindent\raggedright\textbf{PVLDB Artifact Availability:}\\
The source code, data, and/or other artifacts have been made available at \url{\vldbavailabilityurl}.
\endgroup
}

\section{Introduction}
\label{introduction}

AutoML is the process by which machine learning models are built automatically for a new dataset. Given a dataset, AutoML systems perform a search over valid data transformations and learners, along with hyper-parameter optimization for each learner~\cite{VolcanoML}. Choosing the transformations and learners over which to search is our focus.
A significant number of systems mine from prior runs of pipelines over a set of datasets to choose transformers and learners that are effective with different types of datasets (e.g. \cite{NEURIPS2018_b59a51a3}, \cite{10.14778/3415478.3415542}, \cite{autosklearn}). Thus, they build a database by actually running different pipelines with a diverse set of datasets to estimate the accuracy of potential pipelines. Hence, they can be used to effectively reduce the search space. A new dataset, based on a set of features (meta-features) is then matched to this database to find the most plausible candidates for both learner selection and hyper-parameter tuning. This process of choosing starting points in the search space is called meta-learning for the cold start problem.  

Other meta-learning approaches include mining existing data science code and their associated datasets to learn from human expertise. The AL~\cite{al} system mined existing Kaggle notebooks using dynamic analysis, i.e., actually running the scripts, and showed that such a system has promise.  However, this meta-learning approach does not scale because it is onerous to execute a large number of pipeline scripts on datasets, preprocessing datasets is never trivial, and older scripts cease to run at all as software evolves. It is not surprising that AL therefore performed dynamic analysis on just nine datasets.

Our system, {\sysname}, provides a scalable meta-learning approach to leverage human expertise, using static analysis to mine pipelines from large repositories of scripts. Static analysis has the advantage of scaling to thousands or millions of scripts \cite{graph4code} easily, but lacks the performance data gathered by dynamic analysis. The {\sysname} meta-learning approach guides the learning process by a scalable dataset similarity search, based on dataset embeddings, to find the most similar datasets and the semantics of ML pipelines applied on them.  Many existing systems, such as Auto-Sklearn \cite{autosklearn} and AL \cite{al}, compute a set of meta-features for each dataset. We developed a deep neural network model to generate embeddings at the granularity of a dataset, e.g., a table or CSV file, to capture similarity at the level of an entire dataset rather than relying on a set of meta-features.
 
Because we use static analysis to capture the semantics of the meta-learning process, we have no mechanism to choose the \textbf{best} pipeline from many seen pipelines, unlike the dynamic execution case where one can rely on runtime to choose the best performing pipeline.  Observing that pipelines are basically workflow graphs, we use graph generator neural models to succinctly capture the statically-observed pipelines for a single dataset. In {\sysname}, we formulate learner selection as a graph generation problem to predict optimized pipelines based on pipelines seen in actual notebooks.

 
{\sysname} does learner and transformation selection, and hence is a component of an AutoML systems. To evaluate this component, we integrated it into two existing AutoML systems, FLAML \cite{flaml} and Auto-Sklearn \cite{autosklearn}.  
We chose FLAML because it does not yet have any meta-learning component for the cold start problem and instead allows user selection of learners and transformers. The authors of FLAML explicitly pointed to the fact that FLAML might benefit from a meta-learning component and pointed to it as a possibility for future work. For FLAML, if mining historical pipelines provides an advantage, we should improve its performance. We also picked Auto-Sklearn as it does have a learner selection component based on meta-features, as described earlier~\cite{autosklearn2}. For Auto-Sklearn, we should at least match performance if our static mining of pipelines can match their extensive database. For context, we also compared {\sysname} with the recent VolcanoML~\cite{VolcanoML}, which provides an efficient decomposition and execution strategy for the AutoML search space. In contrast, {\sysname} prunes the search space using our meta-learning model to perform hyperparameter optimization only for the most promising candidates. 

The contributions of this paper are the following:
\begin{itemize}
    \item Section ~\ref{sec:mining} defines a scalable meta-learning approach based on representation learning of mined ML pipeline semantics and datasets for over 100 datasets and ~11K Python scripts.  
    \newline
    \item Sections~\ref{sec:kgpipGen} formulates AutoML pipeline generation as a graph generation problem. {\sysname} predicts efficiently an optimized ML pipeline for an unseen dataset based on our meta-learning model.  To the best of our knowledge, {\sysname} is the first approach to formulate  AutoML pipeline generation in such a way.
    \newline
    \item Section~\ref{sec:eval} presents a comprehensive evaluation using a large collection of 121 datasets from major AutoML benchmarks and Kaggle. Our experimental results show that {\sysname} outperforms all existing AutoML systems and achieves state-of-the-art results on the majority of these datasets. {\sysname} significantly improves the performance of both FLAML and Auto-Sklearn in classification and regression tasks. We also outperformed AL in 75 out of 77 datasets and VolcanoML in 75  out of 121 datasets, including 44 datasets used only by VolcanoML~\cite{VolcanoML}.  On average, {\sysname} achieves scores that are statistically better than the means of all other systems. 
\end{itemize}


\section{Related Work}
\label{related_work}

In this section, we summarize the related work and restrict our review to  meta-learning approaches for AutoML, dataset embeddings, and processing tabular structured data. 

\subsubsection*{Learner and preprocessing selection} 
 In most AutoML systems, learner and pre-processing selection for the cold start problem is driven by a database of actual executions of pipelines and data; e.g., \cite{al}, \cite{autosklearn}, \cite{NEURIPS2018_b59a51a3}.  This database often drives both learner selection and hyper parameter optimization (HPO), so we focus here more on how the database is collected or applied to either problem, since the actual application to learner selection or HPO is less relevant.  For HPO, some have cast the application of the database as a multi-task problem (see \cite{multitaskBO}), where the hyperparameters for cold start are chosen based on multiple related datasets. Others, for instance, \cite{autosklearn,Reif2012}, compute a database of dataset meta-features on a variety of OpenML \cite{OpenML} datasets, including dataset properties such as the number of numerical attributes, the number of samples or skewness of the features in each dataset.  
 
These systems measure similarity between datasets and use pipelines from the nearest datasets based on the distance between the datasets' feature vectors as we do, but the computation of these vectors is different, as we describe in detail below.  Auto-Sklearn 2.0 \cite{autosklearn2} defines instead a static portfolio of pipelines that work across a wide variety of datasets, and use these to cold-start the learner selection component - that is, every new dataset uses the same set of pipelines.  Others have created large matrices documenting the performance of candidate pipelines for different datasets and viewed the selection of related pipelines as a collaborative filtering problem \cite{NEURIPS2018_b59a51a3}.
 
\subsubsection*{Dataset embeddings}
The most used mechanism to capture dataset features rely on the use of meta-features for a dataset such as \cite{autosklearn,Reif2012}.  These dataset properties vary from simple, such as number of classes (see, e.g. \cite{Engels98usinga}), to complex and expensive, such as statistical features (see, e.g. \cite{Vilalta_usingmeta-learning}) or landmark features (see, e.g. \cite{Pfahringer00meta-learningby}).  As pointed out in Auto-Sklearn 2.0 \cite{autosklearn2}, these meta-features are not defined with respect to certain column types such as categorical columns, and they are also expensive to compute, within limited budgets.  The dataset embedding we adopt is builds individual column embeddings, and then pools these for a table level embedding.  Similar to our approach, \citet{drori2019automl} use pretrained language models to get dataset embeddings based on available dataset textual information, e.g. title, description and keywords. Given these embeddings, their approach tries to find the most similar datasets and their associated baselines. Unlike \cite{drori2019automl}, our approach relies on embedding the actual data inside the dataset and not just their overall textual description, which in many cases is not available. OBOE \cite{Yang_2019} uses the performance of a few inexpensive, informative models to compute features of a model.

\subsubsection*{Pipeline generation}
There is a significant amount of work viewing the selection of learners as well as hyperparameters as a bayesian optimization problem like \cite{multitaskBO,autoweka}.  Other systems have used evolutionary algorithms along with user defined templates or grammars for this purpose such as TPOT \cite{le2020scaling} or Recipe \cite{S2017RECIPEAG}.  Still, others have viewed the problem of pipeline generation as a probabilistic matrix factorization \cite{NEURIPS2018_b59a51a3}, an AI planning problem when combined with a user specified grammar \cite{ICAPS20paper208,Ml-plan}, a bayesian optimization problem combined with Monte Carlo Tree Search \cite{ijcai2019-457}, or an iterative alternating direction method of multipliers optimization (ADMM) problem \cite{liu2019admm}.  Systems like VolcanoML focus on an efficient decomposition of the search space ~\cite{VolcanoML}. To the best of our knowledge, 
{\sysname} is the first system to cast the actual generation of pipelines as a neural graph generation problem.

Some recent AutoML systems have moved away from the fairly linear pipelines generated by most earlier systems to use ensembles or stacking extensively.  H2O for instance uses fast random search in combination with ensembling for the problem of generating pipelines \cite{LeDell2020H2OAS}.  Others rely on "stacking a bespoke set of models in a predefined order", where stacking and training is handled in a special manner to achieve strong performance \cite{erickson2020autogluontabular}. Similarly, PIPER \cite{piper} uses a greedy best-first search algorithm to traverse the space of partial pipelines guided over a grammar that defines complex pipelines such as Directed Acyclic Graphs (DAGs).  The pipelines produced by PIPER are more complex than the linear structures used in the current AutoML systems we use to test our ideas for historical pipeline modeling, and we do not use ensembling techniques yet in our approach.
Neither is precluded, however, because {\sysname} meta-learning model can generate any type of structures, including complex structures that mined pipelines may have.

\section{The {\sysname} Scalable Meta-Learning}
\label{sec:mining}

\begin{figure}
\ncp\ncp
 \centering
   \includegraphics[width=0.9\columnwidth]{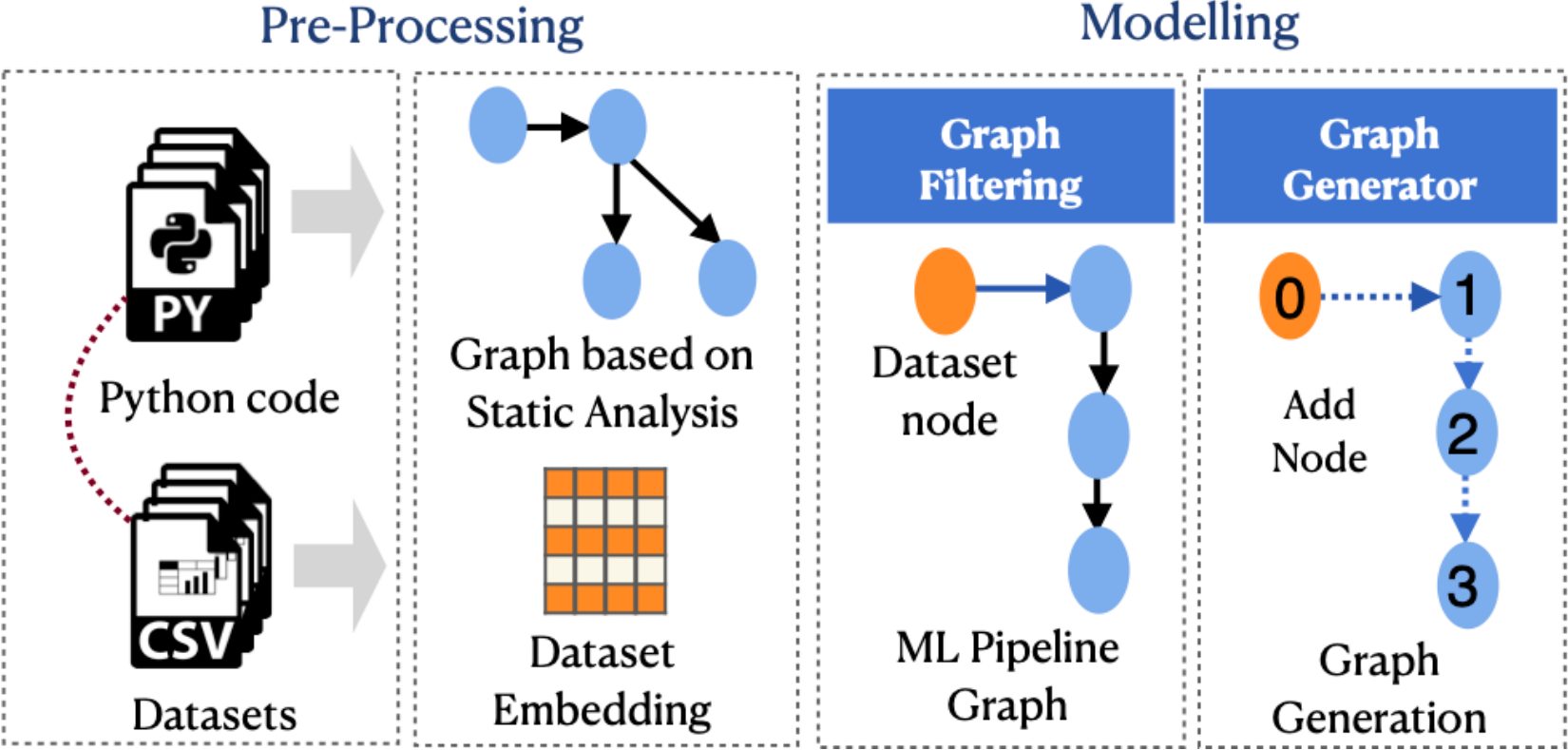} 
  \ncp\ncp\ncp\ncp\ncp\ncp
  \caption{An overview of {\sysname}'s meta-learning approach for mining a database of ML pipelines to train a graph generator model to predict ML pipeline skeletons in the form of graphs.}
  \ncp\ncp\ncp\ncp\ncp\ncp
  \label{fig:mining}
\end{figure}

Our meta-learning approach is based on mining large databases of ML pipelines associated with the used datasets, as illustrated in Figure~\ref{fig:mining}. 
The mining process uses static program analysis instead of executing the actual pipeline scripts or preparing the actual raw data. The {\sysname} meta-learning component enhances the search strategy of existing AutoML systems, such as AutoSklearn and FLAML, and allows these systems to handle ad-hoc datasets, i.e., unseen ones. 
To retain a maximal degree of flexibility, {\sysname} captures metadata and semantics in a flexible graph format, and relies on graph generator models as the database of pipelines. 

Unlike existing meta-learning approaches, our approach is designed to learn from a large scale database and achieve high degree of coverage and diversity. Several ML portals, such as Kaggle or OpenML~\cite{OpenML}, provide access to thousands of datasets associated with hundreds of thousands of public notebooks, i.e., ML pipelines/code. 
{\sysname} mines these large databases of datasets and pipelines using static analysis and filters them into ML pipelines customized for the learner selection problem.
The {\sysname} meta-learning approach leverages~\cite{graph4code} for code understanding via static analysis of scripts/code of ML pipelines. It extracts the semantics of these scripts as code and form an initial graph for each script. 

{\sysname} cleans the graphs generated by ~\cite{graph4code} to keep the semantic required for the ML meta-learning process. Furthermore, our approach introduces dataset nodes and interlinks the relevant pipeline semantic to them. So, our meta-learning approach produces MetaPip, a highly interconnected graph of seen datasets and pipelines applied to them. We also developed a deep embedding model to find the closest datasets to an unseen one, i.e., to effectively prune MetaPip. We then train a deep graph generator model~\cite{deepgmg} using MetaPip. This model is the core of our meta-learning component as illustrated in Figure~\ref{fig:mining} and discussed in the next section.

\subsection{Graph Representation of Code Semantics}
\label{sec:static}

Static and dynamic program analysis techniques could be used to abstract the semantics of programs and extract language-independent representations of code. A program source code is examined in the static analysis without running the program. In contrast, dynamic analysis examines the source code during runtime to collect memory traces and more detailed statistics specific to the analysis technique. Unlike static analysis, dynamic analysis helps in capturing more rich semantics from programs with the high cost of execution and storing massive memory traces. ML portals, such as Kaggle, have hundreds of thousands of ML pipelines with no instructions for running or managing the environments of these pipelines. {\sysname} combines dataset embedding with static code analysis tools, such as {\sysGC}~\cite{graph4code}, to enrich the collected semantics of ML pipelines while avoiding the need to run them.  

\begin{figure}
\ncp\ncp
 \centering
  \includegraphics[width=0.7\columnwidth]{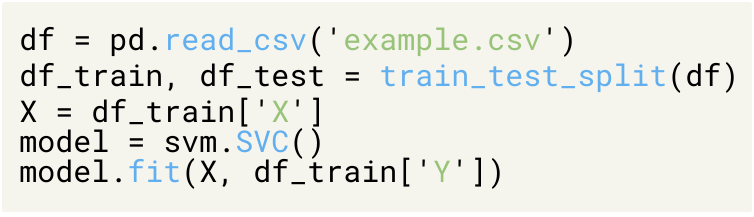}
  \ncp\ncp\ncp\ncp\ncp
  \caption{An example from a data science notebook. 
  }
  \label{running_ex}
  \ncp\ncp\ncp\ncp\ncp\ncp
\end{figure}

\begin{figure}
\ncp\ncp
 \centering
  \includegraphics[width=.45\columnwidth]{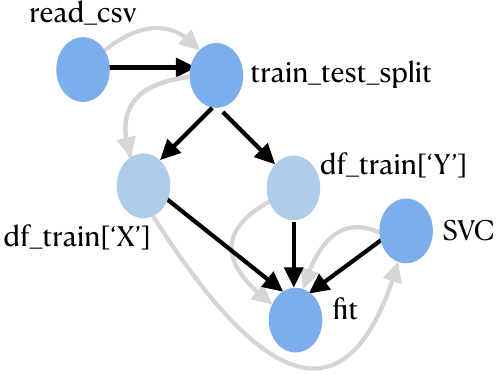}
  \ncp\ncp\ncp\ncp\ncp\ncp\ncp\ncp
  \caption{Code graph corresponding to Figure \ref{running_ex} obtained with {\sysGC}.  The graph shows control flow with gray edges and data flow with black edges. Numerous other nodes and edges are not shown for simplicity.}
  \label{static_analysis}
  \ncp\ncp\ncp\ncp\ncp\ncp\ncp
\end{figure}

{\sysGC} is optimized to efficiently process millions of Python programs, performing interprocedural data flow and control flow analysis to examine for instance, what happens to data that is read from a Pandas dataframe, how it gets manipulated and transformed, and what transformers or estimators get called on the dataframe.  {\sysGC}'s graphs make it explicit what APIs and functions are invoked on objects without the need to model the used  libraries themselves; hence {\sysGC} can scale static analysis to millions of programs. 
Figures~\ref{running_ex} and \ref{static_analysis} show a small code snippet and its corresponding static analysis graph from {\sysGC}, respectively. As shown in Figure \ref{static_analysis}, the graph captures control flow (gray edges), data flow (black edges), as well as numerous other nodes and edges that are not shown in the figure. Examples of these nodes and edges include those capturing location of calls inside a script file and function call parameters. For example, {\sysGC} generates a graph of roughly 1600 nodes and 3700 edges for a Kaggle ML pipeline script of 72 lines of code. The number of nodes and edges dominate the complexity of training a graph generator model. 

\begin{figure}
\ncp
 \centering
  \includegraphics[width=0.55\columnwidth]{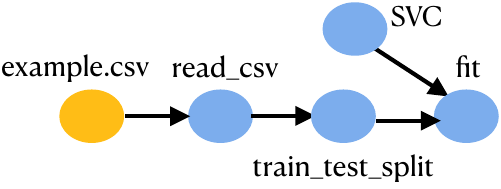}
  \ncp\ncp\ncp\ncp\ncp\ncp
  \caption{Our {\GML} graph of the graph from Figure \ref{static_analysis}, where the abstracted ML pipeline is linked to a dataset node (highlighted in Orange). {\GML} contains at least 96\% less nodes and edges than the original graph while enhancing the overall quality of the graph generation process, as experimented in Section~\ref{sec:study}.  
  }
  \ncp\ncp\ncp\ncp\ncp\ncp\ncp
  \label{abstraction_figure}
\end{figure}

\subsection{MetaPip: from Code to Pipeline Semantics}
\label{abstraction}

For AutoML systems, a pipeline is a set of data transformations, learner selection, and hyper-parameter optimization for each model that is selected. Mined data science notebooks often contain data analysis, data visualization, and model evaluation. Moreover, each notebook is associated with one or more datasets. Thus, it is essential for our meta-learning model to distinguish between different types of pipelines and realize this association with datasets. Existing systems for static code analysis extract general semantics of code and cannot link pipeline scripts to the used datasets.  Thus, the generated graphs by systems, such as {\sysGC}, are scattered and unlinked, i.e., a graph per an ML pipeline script. Moreover, each graph will have nodes and edges that are not relevant for the meta-learning process. \hitext{These irrelevant nodes and edges, i.e., triples, will add noise to the training data. Hence, a meta-learning model will not be able to learn from the abstracted graph pipelines generated by such tools, as shown in Table~\ref{tab:nonabstracted}. We developed a method to filter out this kind of triples from {\sysGC}'s graph and analyze ML pipelines to prepare a training set interconnecting repositories of ML pipeline scripts with their associated datasets. Moreover, our method cleans the noisy nodes and edges and calls to modules outside the target ML libraries. For example, our method will extract triples related to libraries, such as Scikit-learn, XGBoost, and LGBM. These libraries are the most popular among the top-scoring ML pipelines in ML portals. The code for the cleaning method is available at the {\sysname}'s repository.}

Our meta-learning component aims to pick learners and transformer for unseen datasets. Thus, {\sysname} links the filtered ML pipelines with the used datasets. The result of adding these dataset nodes is a highly interconnected graph for ML pipelines, we refer to it as \textit{{\GML}}. Our  {\GML} graph captures both the code and data aspects of ML pipelines. Hence, we can populate the {\GML} graph with datasets from different sources, such as OpenML and Kaggle, and pipelines applied on these datasets. Figure \ref{abstraction_figure} shows the {\GML} graph corresponding to the code snippet in Figure~\ref{running_ex}. {\sysname} utilizes {\GML} to train a model based on a large set of pipelines associated with similar datasets. For example, a \textit{pandas.read\_csv} node will be linked to the used table node, i.e., csv file. In some cases, the code, which reads a csv file, does not explicitly mention the dataset name. The pipelines are usually associated with datasets, such as Kaggle pipelines and datasets, as shown in Figure~\ref{fig:mining}.

\subsection{Dataset Representation Learning}
Our approach efficiently guides the meta-learning process by linking the extracted semantics of pipelines to dataset nodes representing the used datasets. There is a sheer amount of datasets of variable sizes and we need to develop a scalable method for finding the most similar datasets for an unseen one. The pairwise comparison based of the actual content of datasets, i.e, tuples in CSV files, does not scale. Thus, we developed a dataset representation learning method to generate a fixed-size and dense embedding at the granularity of a dataset, e.g., a table or CSV file. The embedding of a dataset $\mathcal{D}$ is the average of its column embeddings, i.e.:
\begin{equation}
\ncp\ncp
\label{equation_dataset_embedding}
    h_{\theta}(\mathcal{D}) = \frac{1}{|\mathcal{D}|} \sum_{c \in \mathcal{D}}{h_\theta(c)}
\end{equation}
where $|\mathcal{D}|$ is the number of columns in $\mathcal{D}$. Our work generalizes the approach outlined in ~\cite{Mueller2019RecognizingVF} for individual column embeddings, where column embeddings are obtained by training a neural network on a binary classification task.  The model learns when two columns represent the same concept, but with different values, as opposed to columns representing different concepts.  Embeddings for an unseen dataset are produced by the last layer of the neural net.

{\sysname} reads datasets only once and leverages PySpark DataFrame to achieve high task and data parallelism.
We use the embeddings of datasets to measure their similarity. With these embeddings, we build an index of vector embeddings for all the datasets in our training set. We utilize efficient libraries~\cite{JDH17} for similarity search of dense vectors to retrieve the most similar dataset to a new input dataset based on its embeddings. Thus, our method scales well and leads to accurate results in capturing similarities between datasets.

\section{The {\sysname} Pipeline Automation}
\label{sec:kgpipGen}

The {\sysname} workflow for pipeline automation is based on our meta-leaning model, as illustrated in Figure~\ref{fig:kgpipGen}. {\sysname} predicts the top-K pipeline skeletons, i.e., a specific set \{$P$, $E$\} of Preprocessor ($P$) and Estimators ($E$), for an unseen dataset ($D$) based on the most similar seen dataset ($SD$), i.e., the nearest neighbour dataset.
{\sysname} starts by finding $SD$ based on the embedding of the unseen dataset. Then, {\sysname} generates the top-K validated ML pipeline graphs $VG$ and converts them into ML pipeline skeletons $\{P,E\}$. Then, it performs hyperparameter optimization using systems, such as  FLAML \cite{flaml} and Auto-Sklearn \cite{autosklearn}, to find the optimum hyperparameters for each pipeline skeleton within a specific time budget.

\subsection{Graph Generation for ML Pipelines}
\begin{figure}
\ncp\ncp
 \centering
   \includegraphics[width=1\columnwidth]{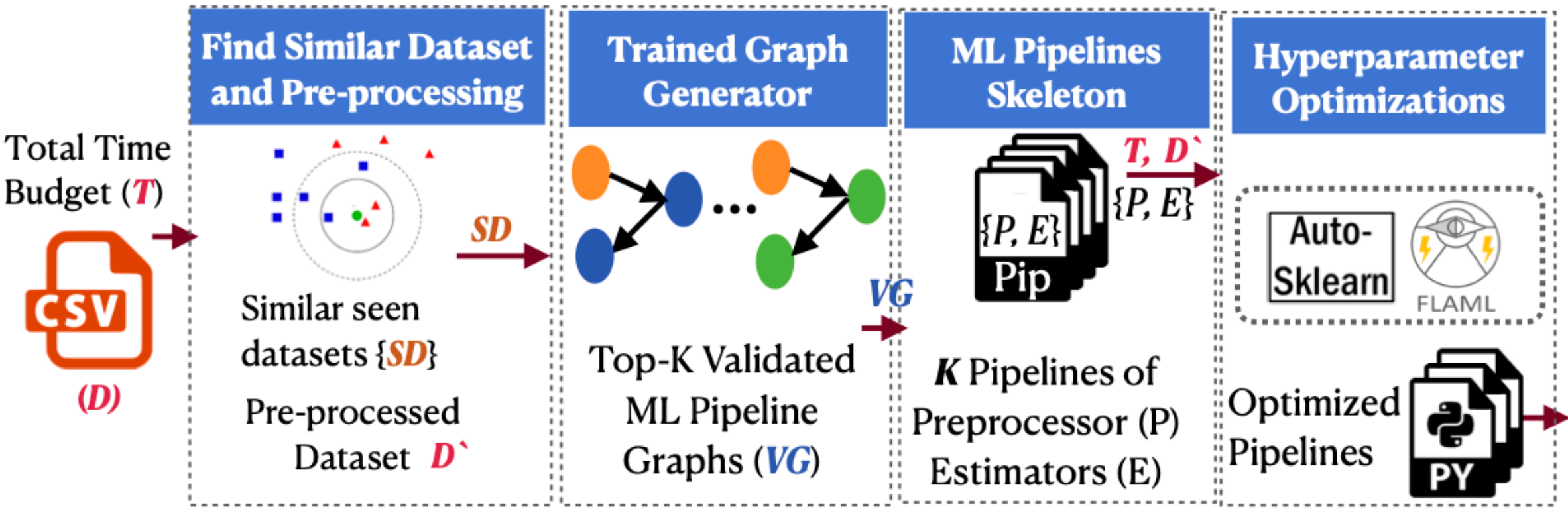} 
  \ncp\ncp\ncp\ncp\ncp\ncp\ncp\ncp\ncp
  \caption{An overview of {\sysname}'s workflows of ML pipeline generation for a given unseen dataset and certain time budget. {\sysname} utilizes systems for hyperparameter optimization, such as FLAML or Auto-Sklearn, to optimize {\sysname}'s top-K predicted pipelines ($VG$), i.e., pruning the search space.}
  \ncp\ncp\ncp\ncp\ncp\ncp\ncp\ncp
  \label{fig:kgpipGen}
\end{figure}

{\sysname} formulates the generation of ML pipelines as a graph generation problem. The intuition behind this idea is that a neural graph generator might capture more succinctly multiple pipelines seen in practice for a given dataset, and might also capture statistical similarities between different pipelines more effectively. To effectively use such a network, we add a single dataset node as the starting point for the filtered pipelines we generate from Python notebooks. The node is assumed to flow into a \texttt{read\_csv} call which is often the starting point for the pipelines. For generating an ML pipeline, we simply pass in a dataset node for the nearest neighbour of the unseen dataset, i.e., the most similar dataset based on content similarity, as shown in Figure~\ref{fig:kgpipGen}.

Our meta-learning model generates ML pipeline graphs in a sequential node-by-node fashion. Algorithm~\ref{algorithm:graph_generation} illustrates the implementation of the graph generation model. For an empty graph $G$ and the most similar dataset $SD$, the algorithm starts by adding an edge between $SD$ and \texttt{pandas.read\_csv}. Then, the graph neural network $f_{AddNode}$ decides whether to add a new node of a certain type. The network $f_{AddEdge}$ decides whether to add an edge to the newly added node. Then, the network $f_{ChooseNode}$ decides the existing node to which the edge is to be added. The While loop at line 7 is repeated repeated until no more edges to be added.  The While loop at line 4 is repeated until no more nodes to be added. The three neural networks, namely $f_{AddNode}$, $f_{AddEdge}$, and  $f_{ChooseNode}$, utilize node embeddings that are learned throughout training via graph propagation rounds. These embeddings capture the structure of ML pipeline graphs.

\begin{algorithm}[t]

\caption{Graph Generation Process}
\label{algorithm:graph_generation}
\small
\SetKwInput{Input}{Input}
\Input{Graph $G$: ($E=\phi$, $V=\phi$), Similar Dataset Node: $SD$, \\ \qquad\quad\   Neural Networks: $f_{AddNode}, f_{AddEdge}, f_{ChooseNode}$}
$V \gets V \cup \{SD, pandas.read\_csv\}$ \\ 
$E \gets E \cup \{(SD, pandas.read\_csv)\}$ \\
$nodeToAdd = f_{AddNode}(V, E)$\\
\While{$nodeToAdd \neq Null$}{
    $V \gets V \cup \{nodeToAdd\}$\\
    $addEdge = f_{AddEdge}(V, E)$ \\
    \While{$addEdge$}{
        $nodeToLink = f_{ChooseNode}(V, E)$\\
        $E \gets E \cup \{(nodeToAdd, nodeToLink)\}$ \\
        $addEdge \gets f_{AddEdge}(V, E)$ \\
    }
    $nodeToAdd \gets f_{AddNode}(V, E)$\\
}
$VG = validate\_pipeline\_graph(G)$ \\
\Return{VG}
\newline
\newline
\ncp\ncp\ncp\ncp\ncp\ncp\ncp\ncp\ncp\ncp\ncp\ncp
\end{algorithm}

The generated graph  $G$ is not guaranteed to be a valid ML pipeline. Thus, \hitext{Algorithm~\ref{algorithm:graph_generation} at line 14} checks that $G$ is a valid ML pipeline graph. In {\sysname}, a graph $G$ is valid if 1) it contains at least one estimator matching the task, i.e., regression or classification, and 2) the estimator is supported by the hyperparameter optimizer (AutoSklearn or FLAML in our case). With these modifications, it is possible to generate ML pipelines for unseen datasets using the closest seen dataset node -- more specifically, its content embedding obtained from the dataset embedding module. We  built \hitext{Algorithm~\ref{algorithm:graph_generation} on top of the system proposed in~\cite{deepgmg}}. This system does not support conditional graph generation at test time by default, i.e., building a graph on top of a provided dataset node. We extended this system to generate valid ML pipeline graphs, as illustrated in \hitext{Algorithm~\ref{algorithm:graph_generation}}.

\subsection{Hyperparameter Optimizion}

\hitext{{\sysname} maps the valid graphs into ML pipeline skeletons, where each skeleton is a set of pre-processors and an estimator with place holders for the optimal parameters. In {\sysname}, the hyperparameter optimizer is responsible for finding the optimal parameters for the pre-processors and learners on the target dataset. Then, {\sysname} replaces the place holders with these parameters. Finally, {\sysname} creates a python script using the pre-processors and estimator achieving the highest scores.}
{\sysname} is well designed to support both  numerical and non-numerical datasets.  Thus, {\sysname}  applies different pre-processing techniques on the given dataset ($D$) and produces a pre-processed dataset ($D'$).
Our pre-processing includes 1) detecting task type (i.e. regression or classification) automatically based on the distribution of the target column 2) automatically inferring accurate data types of columns, 3) \hitext{vectorizing textual columns using word embeddings~\cite{CER2018}}, and 4) imputing missing values in the dataset. In {\sysname}, the hyperparameter optimizer uses $D'$.

Similar to hyperparameter optimizers implemented in AutoML systems, such as FLAML or Auto-Sklearn, {\sysname} works within a provided time budget per dataset. We note here that the majority of the allotted time budget for ML pipeline generation is spent on the hyperparameter optimization; that is, if the user desires only to know what learners would work best for their dataset, {\sysname} can do that almost instantaneously. Given a time budget ($T$), {\sysname} calculates $t$, the time consumed in generating and validating the graphs. {\sysname} then divides the rest of the time budget between the $K$ graphs. Hence, the hyperparameter optimizer has a time limit of ($(T-t) / K$) to optimize each graph independently.   

The hyperparameter optimizer repeatedly applies the learners and pre-processors with different configurations while monitoring the target score metric throughout. {\sysname}  keeps updating its output with the best pipeline skeleton, i.e.,  learners and pre-processors, and its score.  
For example, if the predicted learner is LogisticRegression, it searches for the best combination of regularization type (L1 or L2) and regularization parameter. The difference between hyperparameter optimizers is the search strategy followed to arrive at the best hyperparameters within the allotted time budget. A naive approach would be to perform an exhaustive grid search over all combinations, while a more advanced approach would be to start with promising configurations first.  
We integrate {\sysname} with the hyperparameter optimizers of both FLAML \cite{flaml} and Auto-Sklearn \cite{autosklearn} to demonstrate the generality of {\sysname}. The integration of a hyperparameter optimizer into {\sysname} needs a JSON document of the particular preprocessors and estimators supported by the hyperparameter optimizer. Thus, the integration is relatively easy. Finally, our neural graph generation produces a diverse set of pipelines across runs, allowing for exploration and exploitation.

\ncp\ncp\ncp
\section{Experiments}
\label{sec:eval}

\subsection{Benchmarks} 
We evaluate {\sysname} as well as the other baselines  on four  benchmark datasets: 1) \textit{Open AutoML Benchmark}~\cite{automl_benchmark}, a collection of 39 binary and multi-class \textit{classification} datasets (used by FLAML \cite{flaml}). The datasets are selected such that they are representative of the real world from a diversity of problem domains and of enough difficulty for the learning algorithms. 2) \textit{Penn Machine Learning Benchmark} (PMLB) \cite{pmlb}: Since Open AutoML Benchmark is limited to classification datasets, the authors of FLAML \cite{flaml} evaluated their system on 14 more \textit{regression} datasets selected from PMLB, such that the number of samples is more than 10,000. To demonstrate the generality of our approach, we include those datasets in our evaluation as well. 3) \textit{AL's datasets}: We also evaluate on the datasets used for AL's \cite{al} evaluation which include 6 Kaggle datasets (2 regression and 4 classification) and another 18  classification datasets (9 from PMLB and 9 from OpenML). Unlike other benchmarks, the Kaggle datasets include datasets with textual features. 
4) VolcanoML's datasets: finally, we evaluate {\sysname} on 44 more datasets used by VolcanoML \cite{VolcanoML}. The authors of VolcanoML evaluate their system on a total of 66 datasets from OpenML and Kaggle, from which 11 datasets are not specified, 10 datasets overlap with ours, and 1 dataset consists of image samples. 
\autoref{dataset_stats} includes a summary of all 121 benchmark datasets.
\hitext{The detailed statistics of all datasets are shown in the appendix. 
These statistics include names, number of rows and columns, number of numerical, categorical, and textual features, number of classes, sizes, sources, and papers that evaluated on them.}


\begin{table}[t]
\ncp\ncp
\caption{Breakdown of all 121 datasets used in our evaluation, indicating those used by FLAML$^*$, AL$^\dagger$, and VolcanoML$^\S$.}
\ncp\ncp\ncp
\label{dataset_stats}
 \resizebox{0.8\columnwidth}{!}{%
 
\begin{tabular}{l|llll}
\toprule
                           & \multicolumn{4}{c}{Source} \\
                           \cline{2-5} 
Task                       & AutoML   & PMLB      & OpenML       & Kaggle         \\
\midrule
Binary                     & 22 \small{(18$^*$+1$^{*\dagger}$+3$^{*\S}$)}        &  5 \small{(4\textsuperscript{$\dagger$}+1\textsuperscript{$\dagger\S$})}  &  27 \small{(3\textsuperscript{$\dagger\S$}+3\textsuperscript{$\dagger$}+21$^\S$)}  & 2\textsuperscript{$\dagger$}     \\
Multi-class                & 17 \small{(15$^*$+1$^{*\dagger}$+1$^{*\S}$)}       &  4\textsuperscript{$\dagger$}  &  7 \small{(2\textsuperscript{$\dagger\S$}+1\textsuperscript{$\dagger$} + 4$^\S$})   & 2\textsuperscript{$\dagger$}      \\
Regression                 & 0        & 14$^*$                                                            &  19$^\S$                              & 2\textsuperscript{$\dagger$}     \\
\midrule
Total                      & 39       & 23                                                                               & 53                               & 6   \\
\bottomrule
\end{tabular}
}
\ncp\ncp\ncp\ncp\ncp
\end{table}





\begin{figure*}
\ncp\ncp
\centering
  \includegraphics[width=0.95\textwidth]{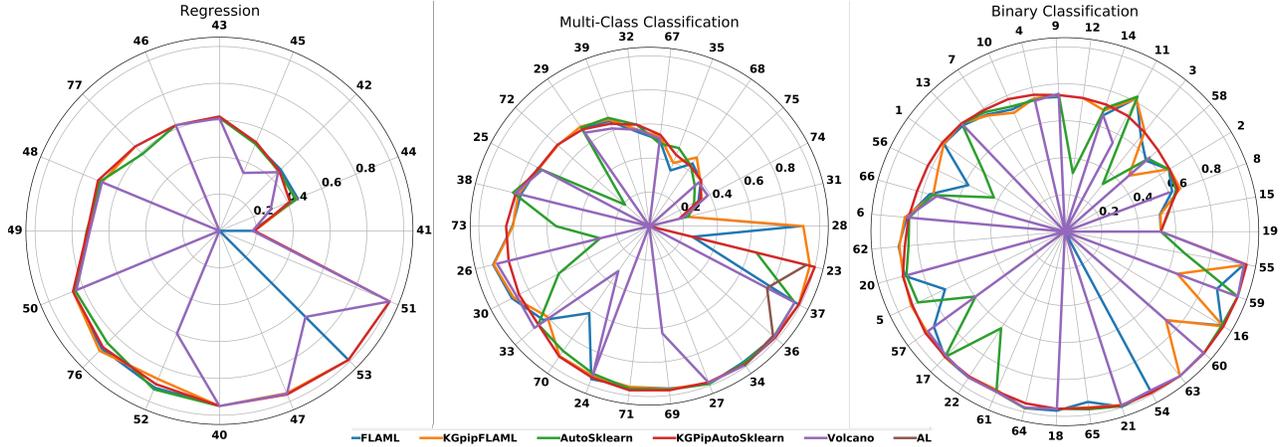}
  \ncp\ncp\ncp\ncp\ncp\ncp\ncp\ncp
  \caption{A radar diagram of the performance of {\sysname} vs. existing systems on multiple tasks (77 datasets) with a time budget of 1 hour for all systems. The outer numbers indicate  different dataset IDs and the ticks inside the figure denote performance ranges of respective metrics; e.g., 0.2, 0.4, ..., etc. for F1 in binary classification. For any dataset, the system with the out most curve has the best performance. As an example, KGpipAutoSklearn and KGpipFLAML achieved 100\% and 97\% F1 on dataset \#23 (multi-class classification) compared to 65\% and 26\% for AutoSklearn and FLAML, respectively. }
  \label{1hr_exps}
  \ncp\ncp\ncp\ncp
\end{figure*}



\subsection{Baselines}
We empirically validate {\sysname} against three AutoML systems: (1) Auto-Sklearn (v0.14.0) \cite{autosklearn} which is the overall winner of multiple challenges in the ChaLearn AutoML competition \cite{chalearn}, and one of the top 4 competitors reported in the Open AutoML Benchmark \cite{automl_benchmark}. (2) FLAML (v0.6.6) \cite{flaml}: an AutoML library designed with both accuracy and computational cost in mind. FLAML outperforms Auto-Sklearn among other systems on two AutoML benchmarks using a low computational budget, (3) AL \cite{al}: a meta-learning-based AutoML approach that utilizes dynamic analysis of Kaggle notebooks, an approach that has similarities to ours, and (4) VolcanoML (v0.5.0)~\cite{VolcanoML}, a recent AutoML approach which proposes efficient decomposition strategies for the large AutoML search spaces. \hitext{In all our experiments, we used the latest code provided by the authors for existing systems, the same exact hardware, time budget, and the parameters recommended by the authors of these systems.}

\subsection{Training Setup} 
Because our approach to mining historical pipelines from scripts is relatively cheap, we can apply it more easily on a wider variety of datasets to form a better base as more and more scripts get generated by domain experts on Kaggle competitions.  \hitext{In this work, we performed program analysis on 11.7K scripts associated with 142 datasets, and then selected those with estimators from \texttt{sklearn}, \texttt{XGBoost} and \texttt{LightGBM} since those were the estimators supported by the most AutoML systems for classification and regression.  This resulted in the selection of 2,046 notebooks for 104 datasets; a vast portion of the 11.7K programs were about exploratory data analysis, or involved libraries that were not supported by Auto-Sklearn \cite{autosklearn} or FLAML (e.g., PyTorch and Keras) \cite{flaml}.} 
We used Macro F1 for classification tasks to account for data imbalance, if any, and use $R^2$ for regression tasks, as in FLAML \cite{flaml}. We  also varied the time budget given to each system between 1 hour and 30 minutes, to measure how fast can {\sysname} find an efficient pipeline compared to other approaches. The time budget is end-to-end, from loading the dataset till producing the best AutoML pipeline. In all experiments, we report averages over 3 runs.

\subsection{Comparison with Existing Systems}
\label{baselines_comparison}



In this section, we evaluate {\sysname} against state-of-the-art approaches; FLAML \cite{flaml} and Auto-Sklearn \cite{autosklearn}. Figure \ref{1hr_exps} shows a radar graph of all systems when given a time budget of 1 hour. It shows the performance of all systems on the three tasks in all benchmarks, namely, binary classification, multi-class classification, and regression. For every dataset, the figure shows  the actual performance metric (F1 for classification and $R^2$ for regression) obtained from every system
 \footnote{The detailed scores for every system and dataset as well as the corresponding names of datasets are shown in tables \ref{tab:dataset_stats}-\ref{tab:detailed_scores_volcano} in the appendix.}. 
Therefore, the out most curve from the center of the radar graph has the best performance. In Figure \ref{1hr_exps}, both variations of {\sysname} achieve the best performance across all tasks, outperforming both FLAML and Auto-Sklearn. We also performed a \textit{two-tailed t-Test} between the performance obtained by {\sysname} compared to the other systems. The results show that {\sysname} achieves  significantly better performance than both FLAML and Auto-Sklearn with a t-Test value of $0.01$ and $0.0002$, respectively (both have $p < 0.05$). 

\begin{table}[t]
\ncp\ncp
\caption{Average scores (mean and standard deviation) of {\sysname} compared to FLAML, Auto-Sklearn, and VolcanoML for binary classification (F1), multi-class classification (F1) and regression ($R^2$) tasks on 77 benchmark datasets. T-test values are for {\sysname} vs. FLAML and {\sysname} vs. Auto-Sklearn.}
\label{averages_1h}
\ncp\ncp\ncp\ncp\ncp\ncp
 \resizebox{1\columnwidth}{!}{%
\begin{tabular}{lllll}
\toprule
                   & Binary & Multi-class   & Regression  & T-Test  \\
\midrule

FLAML              & 0.74 \small{(0.23)}             & 0.70 \small{(0.29)}                  & 0.65 \small{(0.29)} & 0.0129              \\
KGpipFLAML       & 0.81 \small{(0.14)}             & \bf 0.76 \small{(0.24)}                  & \bf 0.72 \small{(0.24)} & -                   \\
Auto-Sklearn        & 0.76 \small{(0.20)}             & 0.65 \small{(0.29)}                  & 0.71 \small{(0.24)} & 0.0002              \\
KGpipAutoSklearn & \bf 0.83 \small{(0.14)}             & 0.73 \small{(0.28)}                  & \bf 0.72 \small{(0.24)} & -     \\ 
VolcanoML &  0.55 \small{(0.43)}  & 0.51 \small{(0.38)} & 0.56 \small{(0.32)} & -     \\ 
\bottomrule
\end{tabular}
}
\ncp\ncp\ncp\ncp\ncp
\end{table}

Table ~\ref{averages_1h} also shows the average F1 and $R^2$ values for classification and regression tasks, respectively. The results show that both variations of {\sysname} achieve better performance compared to both FLAML and Auto-Sklearn over all tasks and datasets. 

\textit{Scalability of {\sysname}'s meta-learning against existing systems}: 
The AL meta-learning approach~\cite{al} mines pipelines using dynamic code analysis, which has high cost as discussed in Section~\ref{sec:static}. Thus, the authors of AL provided a pre-trained meta-learning model on 500 pipelines and 9 datasets, which does not scale to cover various cases. In contrast, we trained our meta-learning model using 2000 pipelines and 142 datasets. None of these datasets were included in the 77 datasets used in testing. AL failed in 22 and timed out in 38 datasets. This shows that the {\sysname} meta-learning approach, which is based on pipelines semantics and dataset representation learning, is more effective.
AL failed on many of the datasets during the fitting process. As the figure shows, {\sysname} still outperforms all other approaches, including AL, significantly. On these datasets, AL achieved the lowest F1 score on binary and multi-class classification tasks with values of 0.36 and 0.36, respectively. This compares to 0.74 and 0.75 by FLAML, 0.73 and 0.68 by Auto-Sklearn,  0.79 and  0.79 by {\sysname FLAML}, and 0.79 and 0.74 by {\sysname Auto-Sklearn}.

\begin{figure}
\ncp\ncp
\centering
  \includegraphics[width=0.45\textwidth]{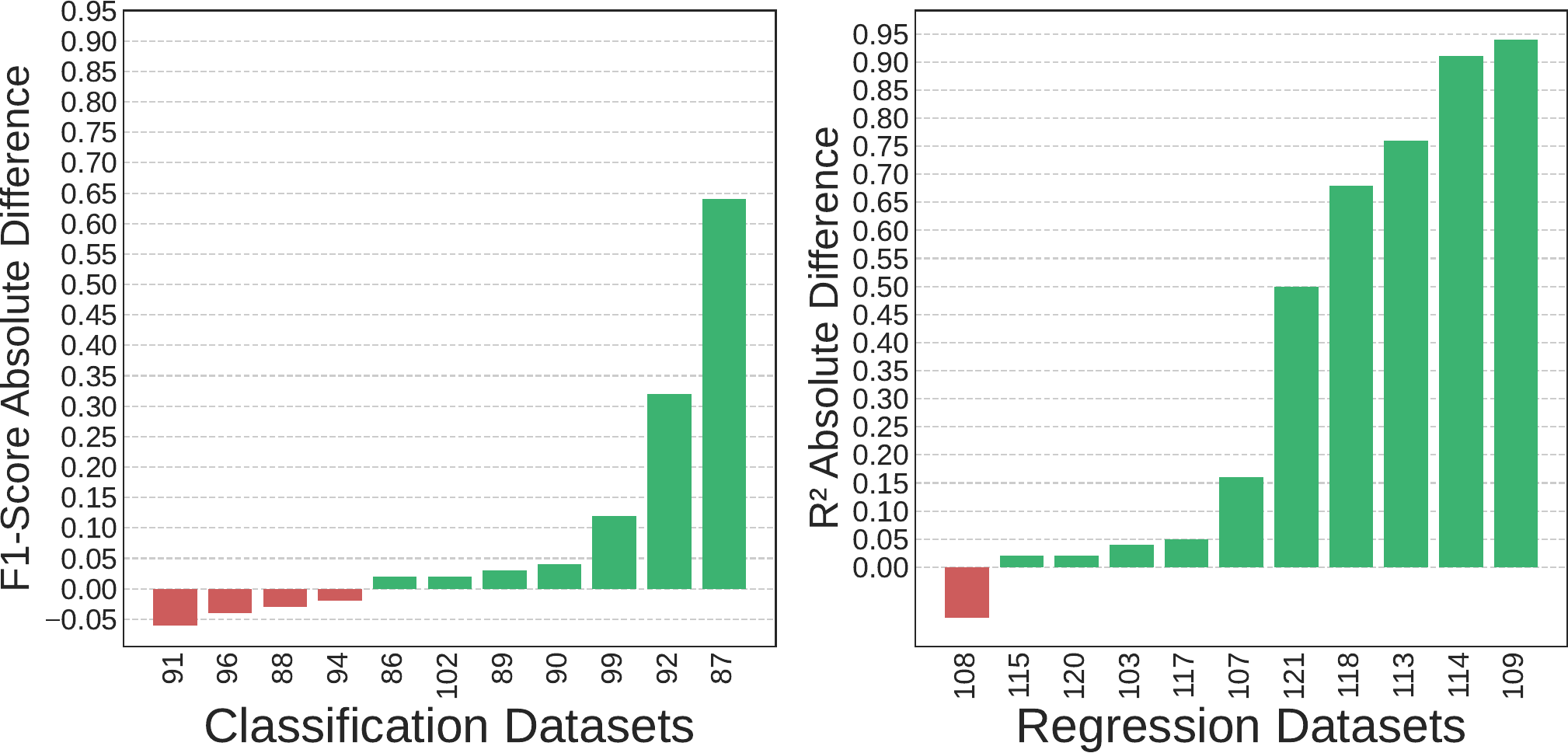}
  \ncp\ncp\ncp\ncp\ncp
  \caption{\hitext{Score difference between KGpipFLAML and VolcanoML on the 44 classification and regression datasets from VolcanoML with a time budget of 1 hour. For brevity, we removed from this Figure the 22 datasets on which both systems perform comparably (within a difference of $\leq 0.01$). 
  }}
  \label{volcanoML_exp}
\end{figure}

\begin{table}[t]
\caption{Average scores (mean and standard deviation) of KGpipFLAML compared to VolcanoML on the 44 datasets from VolcanoML. Overall, KGpipFLAML achieves significantly better compared to VolcanoML, according to a statistical significance test of $p < 0.05$ .}
\label{averages_1h_vs_volcano}
\ncp\ncp\ncp\ncp\ncp\ncp
 \resizebox{1\columnwidth}{!}{%
\begin{tabular}{lllll}
\toprule
                   & Binary & Multi-class   & Regression  & T-Test  \\
\midrule

KGpipFLAML       & \bf 0.82 \small{(0.14)}     & \bf 0.86 \small{(0.16)}   & \bf 0.83 \small{(0.13)} & -                   \\
VolcanoML &  0.69 \small{(0.23)}  & 0.70 \small{(0.31)} & 0.68 \small{(0.25)} & 0.0001     \\ 
\bottomrule
\end{tabular}
}
\ncp\ncp\ncp\ncp
\end{table}

\textit{VolcanoML Datasets}: VolcanoML used a variety of datasets that are not included in our 77 datasets of \autoref{1hr_exps}. Some of these datasets are quite large which are meant to test the the system scalability. Therefore, we also collected all 49 the datasets we could find in their paper and tested the best version of {\sysname} (KGpipFLAML) against VolcanoML on these datasets with a time budget of 1 hour. The performances of KGpipFLAML and VolcanoML are shown in Figure \ref{volcanoML_exp}. For brevity, we omitted from the figure all datasets on which the performance difference between both systems is $\leq 0.01$ \hitext{and the datasets overlapping with the ones shown in Figure \ref{1hr_exps}. On those datasets, KGpipFLAML found a valid pipeline for all of them, sometimes with a decent absolute difference in F1 or $R^2$ scores of  $ \geq 0.90$. Across all the 44 datasets, KGpipFLAML achieved significantly better average of scores compared to VolcanoML (statistical significance test of $p$ < 0.05), see Table~\ref{averages_1h_vs_volcano} for details.}

\begin{figure*}[t]
 \centering
 \ncp\ncp\ncp
 \begin{subfigure}{0.42\textwidth}
    \centering
    \includegraphics[width=0.85\textwidth]{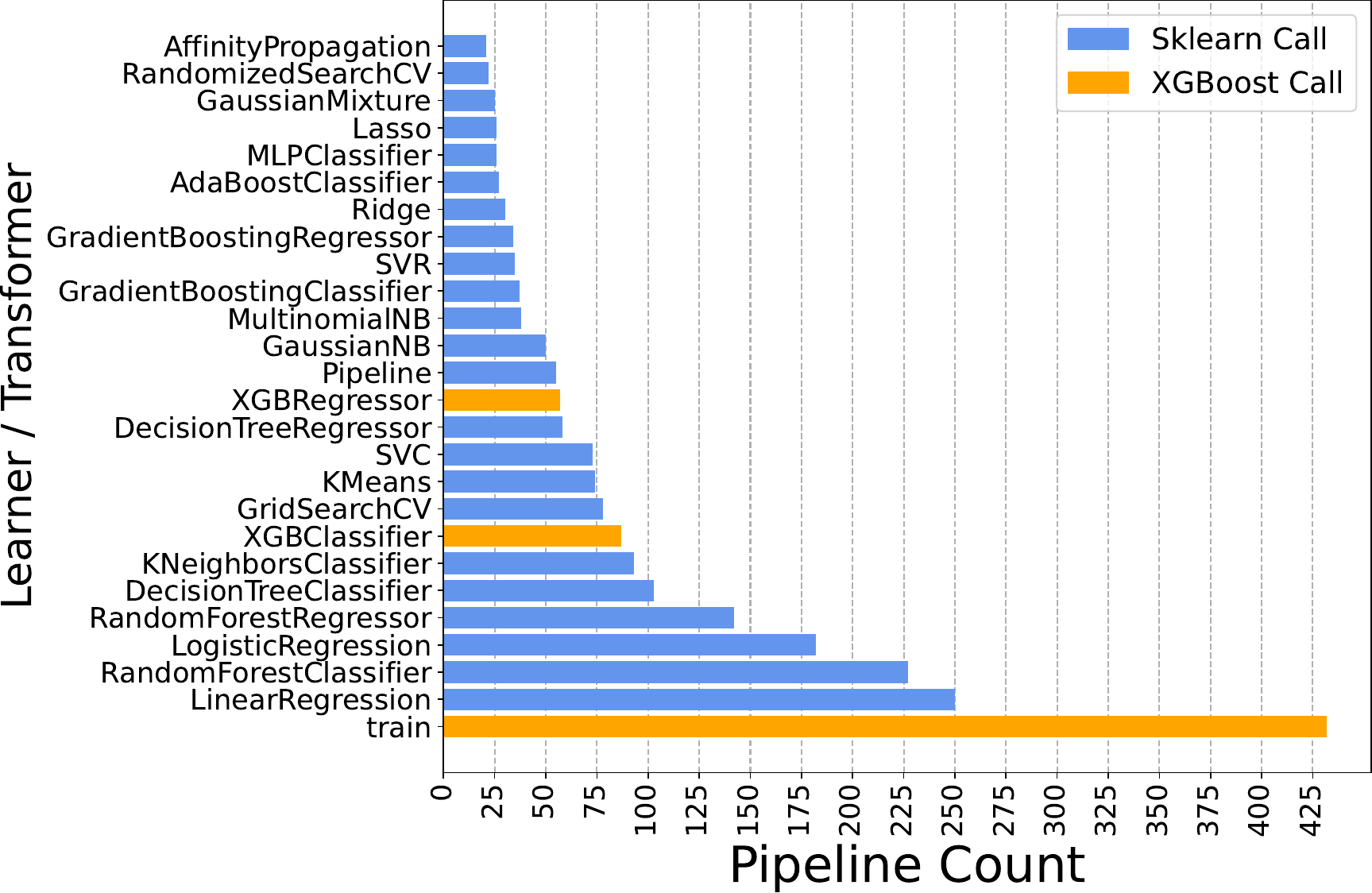}
    \ncp\ncp
    \caption{}
    \label{fig:diversity}
  \end{subfigure}%
  \ncp\ncp\ncp
 \begin{subfigure}{0.42\textwidth}
 \centering
    \includegraphics[width=0.7\textwidth]{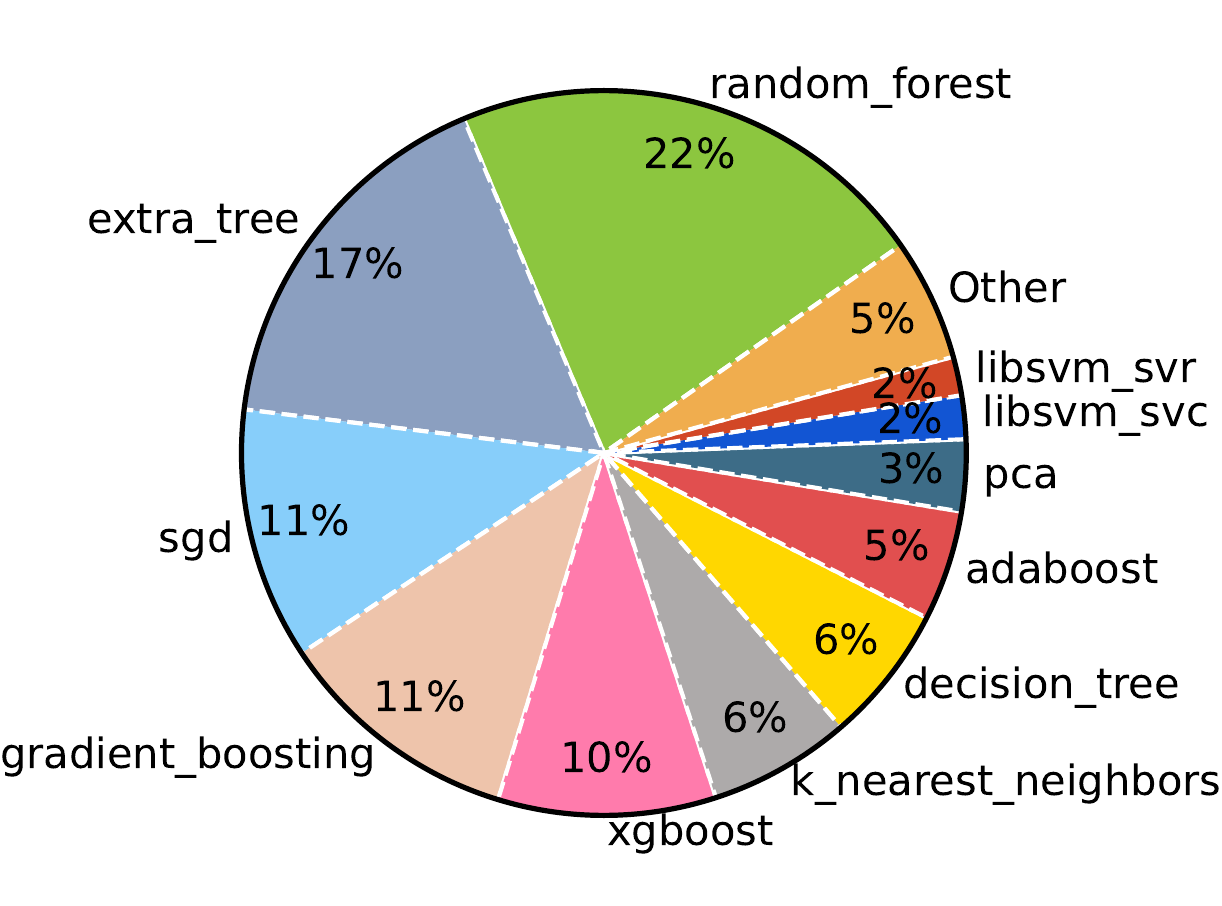}
    \ncp\ncp
    \caption{}
    \label{fig:coverage}
  \end{subfigure}%
  \ncp\ncp\ncp
    \caption{Top learner and transformers selected by {\sysname} (A) are with a wide range of coverage and diversity (B).}
  \label{varying operators chosen}
\end{figure*}

\begin{table}
\small

\centering
\caption{Different aspects comparing a model trained on a set of code graphs vs a model trained on a set of {\GML}  graphs. The model based on original code graphs fails in trivial datasets to generate valid pipelines and limits {\sysname}'s scalability to a larger set of ML pipelines scripts and {\sysname}'s learning by using a fewer number of epochs.}
\ncp\ncp\ncp\ncp
\label{tab:nonabstracted}
 \resizebox{0.85\columnwidth}{!}{%
 
\begin{tabular}{lllll}
\toprule
Dataset/Aspect   & Code Graph & {\GML} Graph   \\
\midrule

kr-vs-kp          & 0 \tiny{(0)}       & 1.00 \tiny{(0)}      \\
nomao             & 0 \tiny{(0)}       & 0.96 \tiny{(0)}      \\
cnae-9            & 0 \tiny{(0)}       & 0.95 \tiny{(0.01)}   \\
mfeat-factors     & 0 \tiny{(0)}       & 0.98 \tiny{(0)}      \\ 
segment           & 0 \tiny{(0)}       & 0.98 \tiny{(0)}      \\
\midrule
Avg. F1           & 0 \tiny{(0)}       & 0.97 \tiny{(0.02)}      \\
No. Nodes           & 29,139        &974       \\
No. Edges           &252,486        &1,052       \\
Training Time            & 175 (min)      & 2 (min) \\
\bottomrule
\end{tabular}
}

\ncp\ncp\ncp\ncp\ncp
\end{table}

\subsection{Ablation Study}
\label{sec:study}

\subsubsection{The effectiveness of {\GML}}
Our {\GML} approach manages to reduce dramatically the number of nodes and edges in the code graph.  Using the original graph obtained from static analysis, it produces the {\GML} graph that focuses on the core aspects needed to train a graph generation model for ML pipelines, such as data transformations, learner selection, and hyper-parameter selection. This experiment investigates the scalability of our graph generation model based on two different training sets, i.e., the sets of {\GML} graphs described in section~\ref{abstraction} vs. the original set of code graphs from static analysis for the same ML pipeline scripts.

For this experiment, we use a small-scale training set of 82 pipeline graphs pertaining to one classification dataset. The original code graphs for these 82 pipelines include 29,139 nodes and 252,486 edges. Our {\GML} graph, however, includes 974 nodes and 1052 edges. This is a graph reduction rate of at least 96.6\%, Figure \ref{tab:nonabstracted} shows these detailed statistics. The main investigation here is whether this huge reduction ratio will help improving the accuracy and scalability of our graph generation model. We train one model on the original code graph and another on the {\GML} graphs. Both models are trained for 15 epochs with the same set of hyperparameters. It is worth noting that due to the huge time required to process the nodes and edges in the code graph, we had to reduce the number of epochs from 400 to 15.

We test the performance of {\sysname} when trained on both graphs on the most trivial binary and multi-class classification datasets in the AutoML benchmark. These are the  datasets where the F1 score of all the reported systems in section \ref{baselines_comparison} is above 0.9. The result is a total of 5 datasets (1 binary and 4 multi-class). Both models use Auto-Sklearn as the hyperparameter optimizer with a time budget of 15 minutes and 3 graphs. We take the average of three runs. The results are summarized in Table \ref{tab:nonabstracted}. For these trivial datasets, the model trained using code graphs did not manage to generate any valid ML pipeline. This means the model failed to capture the core aspects of ML pipelines, i.e., valid transformation or learners. Moreover, our {\GML} approach helps {\sysname} to reduce the training time by 99\%, as shown in Table \ref{tab:nonabstracted}.

\subsubsection{The {\sysname} meta-learning quality}

This experiment tests the quality of our meta-learning component. We test the performance as we vary the number of graphs selected from the graph generation phase before feeding it to the hyper-parameter optimization module.  Table \ref{tab_top_graphs} shows the KGpipFLAML performance as we vary the number of predicted graphs between 3, 5 and 7. 

\begin{table}[t]
\ncp\ncp\ncp
\caption{Performance of KGpipFLAML (mean and standard deviation) as we vary the number of predicted pipeline graphs within 30 minutes time limit. We obtained similar results for KGpipAutoSklearn, and hence omitted its results.}
\ncp\ncp\ncp
\begin{tabular}{llll}
\toprule
             & Binary  & Multi-Class  & Regression \\
\midrule
Top-3 graphs &     0.80 (0.14)                  &       0.70 (0.31)                     &    0.71 (0.23)        \\
Top-5 graphs &       0.81 (0.14)                &     0.73 (0.26)                       &      0.70 (0.23)      \\
Top-7 graphs &      0.81 (0.14)                 &    0.75 (0.24)                        &   0.71 (0.24) \\
\bottomrule
\end{tabular}
\label{tab_top_graphs}
\ncp\ncp\ncp\ncp\ncp\ncp
\end{table}

With only 3 graphs, {\sysname} is still outperforming FLAML (second best system after KGpipFLAML) , although the effect is weaker (t-Test value = 0.06). Compared to Auto-sklearn (third best system after KGpipFLAML), all variations have similar or better performance, but the difference is insignificant. 
This experiment shows that even with three graphs, {\sysname} outperforms FLAML and KGpipAuto-Sklearn, i.e., the correct pipelines often appear in the top 3. As another assessment of the quality of our predictions, we measure where in our ranked list of predicted pipelines the best pipeline turned out to be.  Ideally, the top pipeline would always be first, and we use Mean Reciprocal Rank (MRR) to measure how close to that our predictions are.  Across all runs, the MRR is 0.71, indicating that the top pipeline is typically very near the top.

\subsubsection{The {\sysname} meta-learning diversity} 
One question we addressed is whether {\sysname} produced different pipelines for the \textit{same dataset} across different runs. This gives us a sense of whether {\sysname} is deterministic, or whether it produces different pipelines to help with pruning the AutoML search space.  We took different runs for the exact same dataset, and created a list of learners and transformers produced for each dataset across runs.  The list was limited by the shortest number of learners and transformers produced across runs.  We then computed correlations for datasets across runs 1, 2, and 3.  The correlations ranged from 0.60 - 0.64, suggesting that the runs did not produce the same transformers and learners across runs.
We also examined the types of learners selected by {\sysname} for consideration.  
Figure~\ref{fig:diversity} shows the learners and transformers found at least 20 times in the training pipelines.  One can see from the figure that {\sysname} does not blindly output learners and transformers by counts. 
Figure~\ref{fig:coverage} shows more diversity in what was selected overall. So, a variety of methods are covered by {\sysname}.



\section{Conclusion}
\label{sec:con} 

This paper proposed a novel formulation for the AutoML problem as a graph generation problem, where we can pose learner and pre-processing selection as a generation of different graphs representing ML pipelines. Hence, we developed the {\sysname} system based on mining large repositories of scripts, and leveraging recent techniques for static code analysis. {\sysname} utilized embeddings generated based on dataset contents to predict and optimize a set of ML pipelines based on the most similar seen datasets. {\sysname} is designed to work with AutoML systems, such as Auto-Sklearn and FLAML, to utilize their hyperparameter optimizers. We conducted the most comprehensive evaluation of 121 datasets, including the datasets used by FLAML, VolcanoML, and AL. Our comprehensive evaluation shows that {\sysname} significantly improves the performance of FLAML and Auto-Sklearn in classification and regression tasks. Moreover, {\sysname} outperformed AL, which is based on a more costly meta-learning process, in 97\% of the datasets. This outstanding performance shows that the KGpip meta-learning approach is more effective and efficient. Finally, {\sysname} outperforms VolcanoML in 62\% of the datasets and ties with it in 22\%.



\bibliographystyle{ACM-Reference-Format}
\bibliography{references}

\newpage
\appendix

\section{Statistics of All Datasets}
\begin{table*}
\scriptsize
\centering
\caption{Statistics of the 77 benchmark datasets used in FLAML and AL. From left to right: dataset name, number of rows, number of columns, number of numerical columns, number of categorical columns, number of textual columns, number of classes (for classification datasets), size in MB, source of the dataset, and papers that evaluated on the dataset.}
\begin{tabular}{lllllllllllll}
\toprule
\textbf{ID} & \textbf{Dataset} & \textbf{\#Rows} & \textbf{\#Cols} & \textbf{\#Num} & \textbf{\#Cat} & \textbf{\#Text} & \textbf{\#Classes}  & \textbf{Size (MB)} & \textbf{Task} & \textbf{Source} & \textbf{Paper} \\
\midrule
1   & adult                                       & 48842   & 14     & 2     & 6     & 8      & 0         & 5.7       & binary      & AutoML & FLAML, AL        \\
2   & airlines                                    & 539383  & 7      & 2     & 4     & 3      & 0         & 18.3      & binary      & AutoML & FLAML            \\
3   & albert                                      & 425240  & 78     & 2     & 78    & 0      & 0         & 155.4     & binary      & AutoML & FLAML            \\
4   & Amazon\_employee\_access                    & 32769   & 9      & 2     & 9     & 0      & 0         & 1.9       & binary      & AutoML & FLAML            \\
5   & APSFailure                                  & 76000   & 170    & 2     & 170   & 0      & 0         & 74.8      & binary      & AutoML & FLAML            \\
6   & Australian                                  & 690     & 14     & 2     & 14    & 0      & 0         & 0.0       & binary      & AutoML & FLAML            \\
7   & bank-marketing                              & 45211   & 16     & 2     & 7     & 9      & 0         & 3.5       & binary      & AutoML & FLAML            \\
8   & blood-transfusion-service-center            & 748     & 4      & 2     & 4     & 0      & 0         & 0.0       & binary      & AutoML & FLAML            \\
9   & christine                                   & 5418    & 1636   & 2     & 1636  & 0      & 0         & 31.4      & binary      & AutoML & FLAML            \\
10  & credit-g                                    & 1000    & 20     & 2     & 7     & 13     & 0         & 0.1       & binary      & AutoML & FLAML            \\
11  & guillermo                                   & 20000   & 4296   & 2     & 4296  & 0      & 0         & 424.5     & binary      & AutoML & FLAML            \\
12  & higgs                                       & 98050   & 28     & 2     & 28    & 0      & 0         & 43.3      & binary      & AutoML & FLAML, VolcanoML \\
13  & jasmine                                     & 2984    & 144    & 2     & 144   & 0      & 0         & 1.7       & binary      & AutoML & FLAML            \\
14  & kc1                                         & 2109    & 21     & 2     & 21    & 0      & 0         & 0.1       & binary      & AutoML & FLAML, VolcanoML \\
15  & KDDCup09\_appetency                         & 50000   & 230    & 2     & 192   & 38     & 0         & 32.8      & binary      & AutoML & FLAML            \\
16  & kr-vs-kp                                    & 3196    & 36     & 2     & 0     & 36     & 0         & 0.5       & binary      & AutoML & FLAML            \\
17  & MiniBooNE                                   & 130064  & 50     & 2     & 50    & 0      & 0         & 69.4      & binary      & AutoML & FLAML            \\
18  & nomao                                       & 34465   & 118    & 2     & 118   & 0      & 0         & 19.3      & binary      & AutoML & FLAML            \\
19  & numerai28.6                                 & 96320   & 21     & 2     & 21    & 0      & 0         & 34.9      & binary      & AutoML & FLAML            \\
20  & phoneme                                     & 5404    & 5      & 2     & 5     & 0      & 0         & 0.3       & binary      & AutoML & FLAML, VolcanoML \\
21  & riccardo                                    & 20000   & 4296   & 2     & 4296  & 0      & 0         & 414.0     & binary      & AutoML & FLAML            \\
22  & sylvine                                     & 5124    & 20     & 2     & 20    & 0      & 0         & 0.4       & binary      & AutoML & FLAML            \\
23  & car                                         & 1728    & 6      & 4     & 0     & 6      & 0         & 0.1       & multi-class & AutoML & FLAML            \\
24  & cnae-9                                      & 1080    & 856    & 9     & 856   & 0      & 0         & 1.8       & multi-class & AutoML & FLAML            \\
25  & connect-4                                   & 67557   & 42     & 3     & 42    & 0      & 0         & 5.5       & multi-class & AutoML & FLAML            \\
26  & covertype                                   & 581012  & 54     & 7     & 54    & 0      & 0         & 71.7      & multi-class & AutoML & FLAML, AL        \\
27  & dilbert                                     & 10000   & 2000   & 5     & 2000  & 0      & 0         & 176.0     & multi-class & AutoML & FLAML            \\
28  & dionis                                      & 416188  & 60     & 355   & 60    & 0      & 0         & 110.1     & multi-class & AutoML & FLAML            \\
29  & fabert                                      & 8237    & 800    & 7     & 800   & 0      & 0         & 13.0      & multi-class & AutoML & FLAML            \\
30  & Fashion-MNIST                               & 70000   & 784    & 10    & 784   & 0      & 0         & 148.0     & multi-class & AutoML & FLAML            \\
31  & helena                                      & 65196   & 27     & 100   & 27    & 0      & 0         & 14.6      & multi-class & AutoML & FLAML            \\
32  & jannis                                      & 83733   & 54     & 4     & 54    & 0      & 0         & 36.7      & multi-class & AutoML & FLAML            \\
33  & jungle\_chess\_2pcs\_raw\_endgame\_complete & 44819   & 6      & 3     & 6     & 0      & 0         & 0.6       & multi-class & AutoML & FLAML            \\
34  & mfeat-factors                               & 2000    & 216    & 10    & 216   & 0      & 0         & 1.4       & multi-class & AutoML & FLAML            \\
35  & robert                                      & 10000   & 7200   & 10    & 7200  & 0      & 0         & 268.1     & multi-class & AutoML & FLAML            \\
36  & segment                                     & 2310    & 19     & 7     & 19    & 0      & 0         & 0.3       & multi-class & AutoML & FLAML, VolcanoML \\
37  & shuttle                                     & 58000   & 9      & 7     & 9     & 0      & 0         & 1.5       & multi-class & AutoML & FLAML            \\
38  & vehicle                                     & 846     & 18     & 4     & 18    & 0      & 0         & 0.1       & multi-class & AutoML & FLAML            \\
39  & volkert                                     & 58310   & 180    & 10    & 180   & 0      & 0         & 65.1      & multi-class & AutoML & FLAML            \\
40  & 2dplanes                                    & 40768   & 10     & -     & 10    & 0      & 0         & 2.4       & regression  & PMLB   & FLAML            \\
41  & bng\_breastTumor                            & 116640  & 9      & -     & 9     & 0      & 0         & 6.0       & regression  & PMLB   & FLAML            \\
42  & bng\_echomonths                             & 17496   & 9      & -     & 9     & 0      & 0         & 2.3       & regression  & PMLB   & FLAML            \\
43  & bng\_lowbwt                                 & 31104   & 9      & -     & 9     & 0      & 0         & 2.4       & regression  & PMLB   & FLAML            \\
44  & bng\_pbc                                    & 1000000 & 18     & -     & 18    & 0      & 0         & 220.8     & regression  & PMLB   & FLAML            \\
45  & bng\_pharynx                                & 1000000 & 10     & -     & 10    & 0      & 0         & 68.6      & regression  & PMLB   & FLAML            \\
46  & bng\_pwLinear                               & 177147  & 10     & -     & 10    & 0      & 0         & 10.6      & regression  & PMLB   & FLAML            \\
47  & fried                                       & 40768   & 10     & -     & 10    & 0      & 0         & 8.1       & regression  & PMLB   & FLAML            \\
48  & house\_16H                                  & 22784   & 16     & -     & 16    & 0      & 0         & 5.8       & regression  & PMLB   & FLAML            \\
49  & house\_8L                                   & 22784   & 8      & -     & 8     & 0      & 0         & 2.8       & regression  & PMLB   & FLAML            \\
50  & houses                                      & 20640   & 8      & -     & 8     & 0      & 0         & 1.8       & regression  & PMLB   & FLAML            \\
51  & mv                                          & 40768   & 11     & -     & 11    & 0      & 0         & 5.9       & regression  & PMLB   & FLAML            \\
52  & poker                                       & 1025010 & 10     & -     & 10    & 0      & 0         & 23.0      & regression  & PMLB   & FLAML            \\
53  & pol                                         & 15000   & 48     & -     & 48    & 0      & 0         & 3.0       & regression  & PMLB   & FLAML            \\
54  & breast\_cancer\_wisconsin                   & 569     & 30     & 2     & 30    & 0      & 0         & 0.1       & binary      & PMLB   & AL               \\
55  & detecting-insults-in-social-commentary      & 3947    & 2      & 2     & 0     & 1      & 1         & 0.8       & binary      & Kaggle & AL               \\
56  & fri\_c1\_1000\_25                           & 1000    & 25     & 2     & 25    & 0      & 0         & 0.2       & binary      & OpenML & AL               \\
57  & Hill\_Valley\_with\_noise                   & 1212    & 100    & 2     & 100   & 0      & 0         & 0.8       & binary      & PMLB   & AL               \\
58  & Hill\_Valley\_without\_noise                & 1212    & 100    & 2     & 100   & 0      & 0         & 1.5       & binary      & PMLB   & AL               \\
59  & ionosphere                                  & 351     & 34     & 2     & 34    & 0      & 0         & 0.1       & binary      & PMLB   & AL               \\
60  & MagicTelescope                              & 19020   & 11     & 2     & 11    & 0      & 0         & 1.5       & binary      & OpenML & AL               \\
61  & OVA\_Breast                                 & 1545    & 10936  & 2     & 10936 & 0      & 0         & 103.3     & binary      & OpenML & AL               \\
62  & pc4                                         & 1458    & 37     & 2     & 37    & 0      & 0         & 0.2       & binary      & OpenML & AL, VolcanoML    \\
63  & quake                                       & 2178    & 3      & 2     & 3     & 0      & 0         & 0.0       & binary      & OpenML & AL, VolcanoML    \\
64  & sick                                        & 3772    & 29     & 2     & 7     & 22     & 0         & 0.3       & binary      & OpenML & AL, VolcanoML    \\
65  & spambase                                    & 4601    & 57     & 2     & 57    & 0      & 0         & 1.1       & binary      & PMLB   & AL, VolcanoML    \\
66  & titanic                                     & 891     & 11     & 2     & 6     & 4      & 1         & 0.1       & binary      & Kaggle & AL               \\
67  & car\_evaluation                             & 1728    & 21     & 4     & 21    & 0      & 0         & 0.1       & multi-class & PMLB   & AL               \\
68  & glass                                       & 205     & 9      & 5     & 9     & 0      & 0         & 0.0       & multi-class & PMLB   & AL               \\
69  & kropt                                       & 28056   & 6      & 18    & 3     & 3      & 0         & 0.5       & multi-class & OpenML & AL, VolcanoML    \\
70  & mnist\_784                                  & 70000   & 784    & 10    & 784   & 0      & 0         & 122.0     & multi-class & OpenML & AL, VolcanoML    \\
71  & sentiment-analysis-on-movie-reviews         & 156060  & 3      & 5     & 2     & 0      & 1         & 8.1       & multi-class & Kaggle & AL               \\
72  & splice                                      & 3190    & 61     & 3     & 0     & 61     & 0         & 0.4       & multi-class & OpenML & AL               \\
73  & spooky-author-identification                & 19579   & 2      & 3     & 0     & 1      & 1         & 3.1       & multi-class & Kaggle & AL               \\
74  & wine\_quality\_red                          & 1599    & 11     & 6     & 11    & 0      & 0         & 0.1       & multi-class & PMLB   & AL               \\
75  & wine\_quality\_white                        & 4898    & 11     & 7     & 11    & 0      & 0         & 0.3       & multi-class & PMLB   & AL               \\
76  & housing-prices                              & 1460    & 80     & -     & 37    & 43     & 0         & 0.4       & regression  & Kaggle & AL               \\
77  & mercedes-benz-greener-manufacturing         & 4209    & 377    & -     & 369   & 8      & 0         & 3.1       & regression  & Kaggle & AL               \\

\bottomrule
\end{tabular}

\label{tab:dataset_stats}
\end{table*}

\begin{table*}
\scriptsize
\centering
\caption{Statistics of the 44 datasets used by VolcanoML. From left to right: dataset name, number of rows, number of columns, number of numerical columns, number of categorical columns, number of textual columns, number of classes (for classification datasets), size in MB, source of the dataset, and papers that evaluated on the dataset.}
\begin{tabular}{lllllllllllll}
\toprule
\textbf{ID} & \textbf{Dataset} & \textbf{\#Rows} & \textbf{\#Cols} & \textbf{\#Num} & \textbf{\#Cat} & \textbf{\#Text} & \textbf{\#Classes}  & \textbf{Size (MB)} & \textbf{Task} & \textbf{Source} & \textbf{Paper} \\
\midrule

78  & ailerons                                    & 13750   & 40     & 2     & 40    & 0      & 0         & 2.2       & binary      & OpenML & VolcanoML        \\
79  & analcatdata\_supreme                        & 4052    & 7      & 2     & 7     & 0      & 0         & 0.1       & binary      & OpenML & VolcanoML        \\
80  & bank32nh\_833                               & 8192    & 32     & 2     & 32    & 0      & 0         & 2.1       & binary      & OpenML & VolcanoML        \\
81  & cpu\_act\_761                               & 8192    & 21     & 2     & 21    & 0      & 0         & 0.7       & binary      & OpenML & VolcanoML        \\
82  & cpu\_small\_735                             & 8192    & 12     & 2     & 12    & 0      & 0         & 0.4       & binary      & OpenML & VolcanoML        \\
83  & delta\_ailerons                             & 7129    & 5      & 2     & 5     & 0      & 0         & 0.3       & binary      & OpenML & VolcanoML        \\
84  & delta\_elevators                            & 9517    & 6      & 2     & 6     & 0      & 0         & 0.3       & binary      & OpenML & VolcanoML        \\
85  & eeg-eye-state                               & 14980   & 14     & 2     & 14    & 0      & 0         & 1.6       & binary      & OpenML & VolcanoML        \\
86  & electricity                                 & 45312   & 8      & 2     & 8     & 0      & 0         & 2.9       & binary      & OpenML & VolcanoML        \\
87  & jm1                                         & 10885   & 21     & 2     & 21    & 0      & 0         & 0.8       & binary      & OpenML & VolcanoML        \\
88  & kin8nm\_807                                 & 8192    & 8      & 2     & 8     & 0      & 0         & 0.6       & binary      & OpenML & VolcanoML        \\
89  & mammography                                 & 11183   & 6      & 2     & 6     & 0      & 0         & 0.8       & binary      & OpenML & VolcanoML        \\
90  & mc1                                         & 9466    & 38     & 2     & 38    & 0      & 0         & 1.0       & binary      & OpenML & VolcanoML        \\
91  & ozone-level-8hr                             & 2534    & 72     & 2     & 72    & 0      & 0         & 0.9       & binary      & OpenML & VolcanoML        \\
92  & page-blocks                                 & 5473    & 10     & 2     & 10    & 0      & 0         & 0.2       & binary      & OpenML & VolcanoML        \\
93  & pollen\_871                                 & 3848    & 5      & 2     & 5     & 0      & 0         & 0.1       & binary      & OpenML & VolcanoML        \\
94  & puma32H\_752                                & 8192    & 32     & 2     & 32    & 0      & 0         & 2.3       & binary      & OpenML & VolcanoML        \\
95  & puma8NH\_816                                & 8192    & 8      & 2     & 8     & 0      & 0         & 0.6       & binary      & OpenML & VolcanoML        \\
96  & space\_ga\_737                              & 3107    & 6      & 2     & 6     & 0      & 0         & 0.2       & binary      & OpenML & VolcanoML        \\
97  & waveform-5000                               & 5000    & 40     & 2     & 40    & 0      & 0         & 1.0       & binary      & OpenML & VolcanoML        \\
98  & wind\_847                                   & 6574    & 14     & 2     & 14    & 0      & 0         & 0.4       & binary      & OpenML & VolcanoML        \\
99  & abalone                                     & 4177    & 8      & 28    & 7     & 1      & 0         & 0.2       & multi-class & OpenML & VolcanoML        \\
100 & optdigits                                   & 5620    & 64     & 10    & 64    & 0      & 0         & 0.8       & multi-class & OpenML & VolcanoML        \\
101 & pendigits                                   & 10992   & 16     & 10    & 16    & 0      & 0         & 0.7       & multi-class & OpenML & VolcanoML        \\
102 & satimage                                    & 6430    & 36     & 6     & 36    & 0      & 0         & 2.1       & multi-class & OpenML & VolcanoML        \\
103 & bank32nh\_558                               & 8192    & 32     & -     & 32    & 0      & 0         & 2.4       & regression  & OpenML & VolcanoML        \\
104 & bank8FM                                     & 8192    & 8      & -     & 8     & 0      & 0         & 0.6       & regression  & OpenML & VolcanoML        \\
105 & cpu\_act\_573                               & 8192    & 21     & -     & 21    & 0      & 0         & 1.0       & regression  & OpenML & VolcanoML        \\
106 & cpu\_small\_227                             & 8192    & 12     & -     & 12    & 0      & 0         & 0.6       & regression  & OpenML & VolcanoML        \\
107 & debutanizer                                 & 2394    & 7      & -     & 7     & 0      & 0         & 0.2       & regression  & OpenML & VolcanoML        \\
108 & kin8nm\_189                                 & 8192    & 8      & -     & 8     & 0      & 0         & 1.1       & regression  & OpenML & VolcanoML        \\
109 & Moneyball                                   & 1232    & 14     & -     & 12    & 2      & 0         & 0.1       & regression  & OpenML & VolcanoML        \\
110 & pollen\_529                                 & 3848    & 5      & -     & 5     & 0      & 0         & 0.2       & regression  & OpenML & VolcanoML        \\
111 & puma32H\_308                                & 8192    & 32     & -     & 32    & 0      & 0         & 2.7       & regression  & OpenML & VolcanoML        \\
112 & puma8NH\_225                                & 8192    & 8      & -     & 8     & 0      & 0         & 0.7       & regression  & OpenML & VolcanoML        \\
113 & rainfall\_bangladesh                        & 16755   & 3      & -     & 1     & 2      & 0         & 0.4       & regression  & OpenML & VolcanoML        \\
114 & socmob                                      & 1156    & 5      & -     & 1     & 4      & 0         & 0.1       & regression  & OpenML & VolcanoML        \\
115 & space\_ga\_507                              & 3107    & 6      & -     & 6     & 0      & 0         & 0.5       & regression  & OpenML & VolcanoML        \\
116 & stock                                       & 950     & 9      & -     & 9     & 0      & 0         & 0.1       & regression  & OpenML & VolcanoML        \\
117 & sulfur                                      & 10081   & 6      & -     & 6     & 0      & 0         & 0.6       & regression  & OpenML & VolcanoML        \\
118 & us\_crime                                   & 1994    & 127    & -     & 126   & 1      & 0         & 1.1       & regression  & OpenML & VolcanoML        \\
119 & weather\_izmir                              & 1461    & 9      & -     & 9     & 0      & 0         & 0.1       & regression  & OpenML & VolcanoML        \\
120 & wind\_503                                   & 6574    & 14     & -     & 14    & 0      & 0         & 0.5       & regression  & OpenML & VolcanoML        \\
121 & witmer\_census\_1980                        & 50      & 5      & -     & 4     & 1      & 0         & 0.0       & regression  & OpenML & VolcanoML        \\
\bottomrule
\end{tabular}

\label{tab:dataset_stats_volcano}
\end{table*}

Tables \ref{tab:dataset_stats} and \ref{tab:dataset_stats_volcano} show the statistics of all benchmark datasets including number of rows and columns, number of numerical, categorical, and textual features, number of classes, size, source, and papers that evaluated on these datasets.

\section{Detailed Scores for All Systems}
\begin{table*}
\scriptsize
\centering
\caption{Macro F1 and $R^2$ scores for all systems on all 77 benchmark datasets. The reported scores are averages of 3 runs with 1 hour time budget.}
\begin{tabular}{lllllllll}
\toprule
\textbf{ID} & \textbf{Dataset}                            & \textbf{FLAML} & \textbf{\makecell{KGpip \\ FLAML}} & \textbf{AutoSklearn} & \textbf{\makecell{KGpip \\ AutoSklearn}} & \textbf{VolcanoML} & \textbf{Task} & \textbf{Papers} \\
\midrule
1  & adult                                       & 0.81 & 0.81 & 0.82 & 0.54 & 0.00 & binary      & FLAML, AL        \\
2  & airlines                                    & 0.66 & 0.66 & 0.66 & 0.66 & 0.00 & binary      & FLAML            \\
3  & albert                                      & 0.66 & 0.69 & 0.69 & 0.33 & 0.00 & binary      & FLAML            \\
4  & Amazon\_employee\_access                    & 0.74 & 0.74 & 0.76 & 0.73 & 0.73 & binary      & FLAML            \\
5  & APSFailure                                  & 0.72 & 0.92 & 0.92 & 0.88 & 0.00 & binary      & FLAML            \\
6  & Australian                                  & 0.86 & 0.87 & 0.85 & 0.85 & 0.86 & binary      & FLAML            \\
7  & bank-marketing                              & 0.76 & 0.75 & 0.79 & 0.78 & 0.00 & binary      & FLAML            \\
8  & blood-transfusion-service-center            & 0.62 & 0.67 & 0.65 & 0.64 & 0.60 & binary      & FLAML            \\
9  & christine                                   & 0.73 & 0.74 & 0.74 & 0.75 & 0.75 & binary      & FLAML            \\
10 & credit-g                                    & 0.72 & 0.70 & 0.78 & 0.74 & 0.00 & binary      & FLAML            \\
11 & guillermo                                   & 0.82 & 0.82 & 0.71 & 0.83 & 0.55 & binary      & FLAML            \\
12 & higgs                                       & 0.00 & 0.73 & 0.73 & 0.32 & 0.00 & binary      & FLAML, VolcanoML \\
13 & jasmine                                     & 0.80 & 0.81 & 0.81 & 0.81 & 0.80 & binary      & FLAML            \\
14 & kc1                                         & 0.66 & 0.69 & 0.72 & 0.70 & 0.67 & binary      & FLAML, VolcanoML \\
15 & KDDCup09\_appetency                         & 0.52 & 0.53 & 0.57 & 0.57 & 0.00 & binary      & FLAML            \\
16 & kr-vs-kp                                    & 0.99 & 1.00 & 1.00 & 0.99 & 0.00 & binary      & FLAML            \\
17 & MiniBooNE                                   & 0.94 & 0.94 & 0.94 & 0.94 & 0.93 & binary      & FLAML            \\
18 & nomao                                       & 0.97 & 0.96 & 0.96 & 0.96 & 0.96 & binary      & FLAML            \\
19 & numerai28.6                                 & 0.52 & 0.52 & 0.52 & 0.52 & 0.52 & binary      & FLAML            \\
20 & phoneme                                     & 0.90 & 0.91 & 0.91 & 0.89 & 0.88 & binary      & FLAML, VolcanoML \\
21 & riccardo                                    & 1.00 & 0.99 & 0.99 & 0.99 & 0.99 & binary      & FLAML            \\
22 & sylvine                                     & 0.95 & 0.94 & 0.94 & 0.63 & 0.95 & binary      & FLAML            \\
23 & car                                         & 0.26 & 0.97 & 1.00 & 0.65 & 0.00 & multi-class & FLAML            \\
24 & cnae-9                                      & 0.96 & 0.94 & 0.95 & 0.93 & 0.94 & multi-class & FLAML            \\
25 & connect-4                                   & 0.74 & 0.73 & 0.73 & 0.72 & 0.71 & multi-class & FLAML            \\
26 & covertype                                   & 0.94 & 0.94 & 0.85 & 0.30 & 0.92 & multi-class & FLAML, AL        \\
27 & dilbert                                     & 0.99 & 0.98 & 0.98 & 0.99 & 0.98 & multi-class & FLAML            \\
28 & dionis                                      & 0.88 & 0.90 & 0.00 & 0.00 & 0.00 & multi-class & FLAML            \\
29 & fabert                                      & 0.70 & 0.71 & 0.69 & 0.70 & 0.67 & multi-class & FLAML            \\
30 & Fashion-MNIST                               & 0.91 & 0.90 & 0.86 & 0.60 & 0.88 & multi-class & FLAML            \\
31 & helena                                      & 0.23 & 0.23 & 0.18 & 0.24 & 0.20 & multi-class & FLAML            \\
32 & jannis                                      & 0.56 & 0.57 & 0.60 & 0.60 & 0.57 & multi-class & FLAML            \\
33 & jungle\_chess\_2pcs\_raw\_endgame\_complete & 0.83 & 0.80 & 0.87 & 0.87 & 0.90 & multi-class & FLAML            \\
34 & mfeat-factors                               & 0.97 & 0.98 & 0.99 & 0.98 & 0.99 & multi-class & FLAML            \\
35 & robert                                      & 0.35 & 0.40 & 0.45 & 0.49 & 0.00 & multi-class & FLAML            \\
36 & segment                                     & 0.98 & 0.98 & 0.99 & 0.98 & 0.98 & multi-class & FLAML, VolcanoML \\
37 & shuttle                                     & 0.99 & 0.99 & 0.99 & 0.96 & 0.96 & multi-class & FLAML            \\
38 & vehicle                                     & 0.78 & 0.79 & 0.81 & 0.82 & 0.81 & multi-class & FLAML            \\
39 & volkert                                     & 0.66 & 0.67 & 0.64 & 0.68 & 0.61 & multi-class & FLAML            \\
40 & 2dplanes                                    & 0.95 & 0.95 & 0.95 & 0.95 & 0.95 & regression  & FLAML            \\
41 & bng\_breastTumor                            & 0.18 & 0.19 & 0.19 & 0.18 & 0.18 & regression  & FLAML            \\
42 & bng\_echomonths                             & 0.47 & 0.45 & 0.46 & 0.46 & 0.45 & regression  & FLAML            \\
43 & bng\_lowbwt                                 & 0.62 & 0.62 & 0.62 & 0.61 & 0.61 & regression  & FLAML            \\
44 & bng\_pbc                                    & 0.46 & 0.45 & 0.41 & 0.45 & 0.30 & regression  & FLAML            \\
45 & bng\_pharynx                                & 0.51 & 0.52 & 0.52 & 0.51 & 0.34 & regression  & FLAML            \\
46 & bng\_pwLinear                               & 0.62 & 0.62 & 0.62 & 0.62 & 0.62 & regression  & FLAML            \\
47 & fried                                       & 0.96 & 0.95 & 0.96 & 0.96 & 0.96 & regression  & FLAML            \\
48 & house\_16H                                  & 0.70 & 0.69 & 0.71 & 0.70 & 0.69 & regression  & FLAML            \\
49 & house\_8L                                   & 0.71 & 0.71 & 0.72 & 0.72 & 0.70 & regression  & FLAML            \\
50 & houses                                      & 0.86 & 0.86 & 0.86 & 0.85 & 0.84 & regression  & FLAML            \\
51 & mv                                          & 0.00 & 1.00 & 1.00 & 1.00 & 1.00 & regression  & FLAML            \\
52 & poker                                       & 0.92 & 0.87 & 0.90 & 0.93 & 0.60 & regression  & FLAML            \\
53 & pol                                         & 0.99 & 0.99 & 0.99 & 0.99 & 0.66 & regression  & FLAML            \\
54 & breast\_cancer\_wisconsin                   & 0.98 & 0.99 & 0.99 & 0.99 & 0.98 & binary      & AL               \\
55 & car\_evaluation                             & 0.99 & 1.00 & 1.00 & 0.66 & 0.99 & binary      & AL               \\
56 & detecting-insults-in-social-commentary      & 0.58 & 0.76 & 0.82 & 0.43 & 0.00 & binary      & AL               \\
57 & fri\_c1\_1000\_25                           & 0.88 & 0.92 & 0.93 & 0.60 & 0.92 & binary      & AL               \\
58 & glass                                       & 0.58 & 0.46 & 0.67 & 0.60 & 0.61 & binary      & AL               \\
59 & Hill\_Valley\_with\_noise                   & 0.88 & 0.65 & 1.00 & 1.00 & 1.00 & binary      & AL               \\
60 & Hill\_Valley\_without\_noise                & 0.73 & 0.73 & 1.00 & 1.00 & 1.00 & binary      & AL               \\
61 & ionosphere                                  & 0.94 & 0.93 & 0.94 & 0.94 & 0.94 & binary      & AL               \\
62 & kropt                                       & 0.90 & 0.90 & 0.87 & 0.85 & 0.00 & binary      & AL               \\
63 & MagicTelescope                              & 0.00 & 1.00 & 1.00 & 1.00 & 1.00 & binary      & AL               \\
64 & mnist\_784                                  & 0.98 & 0.98 & 0.95 & 0.98 & 0.98 & binary      & AL               \\
65 & OVA\_Breast                                 & 0.93 & 0.96 & 0.96 & 0.97 & 0.00 & binary      & AL, VolcanoML    \\
66 & pc4                                         & 0.76 & 0.74 & 0.83 & 0.74 & 0.72 & binary      & AL               \\
67 & quake                                       & 0.51 & 0.53 & 0.54 & 0.49 & 0.52 & multi-class & AL               \\
68 & sentiment-analysis-on-movie-reviews         & 0.45 & 0.49 & 0.43 & 0.43 & 0.00 & multi-class & AL               \\
69 & spambase                                    & 0.96 & 0.96 & 0.97 & 0.97 & 0.63 & multi-class & AL               \\
70 & sick                                        & 0.62 & 0.93 & 0.93 & 0.89 & 0.32 & multi-class & AL               \\
71 & splice                                      & 0.95 & 0.95 & 0.97 & 0.96 & 0.00 & multi-class & AL               \\
72 & spooky-author-identification                & 0.00 & 0.72 & 0.72 & 0.19 & 0.00 & multi-class & AL               \\
73 & titanic                                     & 0.80 & 0.80 & 0.84 & 0.55 & 0.00 & multi-class & AL               \\
74 & wine\_quality\_red                          & 0.33 & 0.35 & 0.34 & 0.30 & 0.39 & multi-class & AL               \\
75 & wine\_quality\_white                        & 0.40 & 0.40 & 0.41 & 0.36 & 0.39 & multi-class & AL               \\
76 & housing-prices                              & 0.90 & 0.92 & 0.89 & 0.86 & 0.00 & regression  & AL               \\
77 & mercedes-benz-greener-manufacturing         & 0.59 & 0.65 & 0.65 & 0.59 & 0.00 & regression  & AL    \\
\bottomrule
\end{tabular}

\label{tab:detailed_scores}
\end{table*}

\begin{table*}
\scriptsize
\centering
\caption{Macro F1 and $R^2$ scores for KGpipFLAML and VolcanoML on the 44 datasets used by VolcanoML. The reported scores are averages of 3 runs with 1 hour time budget.}
\begin{tabular}{lllllllll}
\toprule
\textbf{ID} & \textbf{Dataset} & \textbf{\makecell{KGpip \\ FLAML}} & \textbf{VolcanoML} & \textbf{Task} \\
\midrule
78  & ailerons             & 0.89 & 0.88 & binary      \\
79  & analcatdata\_supreme & 0.99 & 0.98 & binary      \\
80  & bank32nh\_833        & 0.79 & 0.79 & binary      \\
81  & cpu\_act\_761        & 0.93 & 0.92 & binary      \\
82  & cpu\_small\_735      & 0.91 & 0.90 & binary      \\
83  & delta\_ailerons      & 0.94 & 0.93 & binary      \\
84  & delta\_elevators     & 0.88 & 0.88 & binary      \\
85  & eeg-eye-state        & 0.96 & 0.97 & binary      \\
86  & electricity          & 0.94 & 0.92 & binary      \\
87  & jm1                  & 0.64 & 0.00 & binary      \\
88  & kin8nm\_807          & 0.87 & 0.90 & binary      \\
89  & mammography          & 0.85 & 0.82 & binary      \\
90  & mc1                  & 0.80 & 0.76 & binary      \\
91  & ozone-level-8hr      & 0.68 & 0.74 & binary      \\
92  & page-blocks          & 0.94 & 0.62 & binary      \\
93  & pollen\_871          & 0.51 & 0.51 & binary      \\
94  & puma32H\_752         & 0.89 & 0.91 & binary      \\
95  & puma8NH\_816         & 0.83 & 0.83 & binary      \\
96  & space\_ga\_737       & 0.83 & 0.87 & binary      \\
97  & waveform-5000        & 0.88 & 0.87 & binary      \\
98  & wind\_847            & 0.86 & 0.86 & binary      \\
99  & abalone              & 0.12 & 0.00 & multi-class \\
100 & optdigits            & 0.99 & 0.99 & multi-class \\
101 & pendigits            & 0.99 & 0.99 & multi-class \\
102 & satimage             & 0.92 & 0.90 & multi-class \\
103 & bank32nh\_558        & 0.60 & 0.56 & regression  \\
104 & bank8FM              & 0.96 & 0.96 & regression  \\
105 & cpu\_act\_573        & 0.98 & 0.98 & regression  \\
106 & cpu\_small\_227      & 0.98 & 0.98 & regression  \\
107 & debutanizer          & 0.83 & 0.67 & regression  \\
108 & kin8nm\_189          & 0.81 & 0.90 & regression  \\
109 & Moneyball            & 0.94 & 0.00 & regression  \\
110 & pollen\_529          & 0.80 & 0.81 & regression  \\
111 & puma32H\_308         & 0.94 & 0.95 & regression  \\
112 & puma8NH\_225         & 0.68 & 0.67 & regression  \\
113 & rainfall\_bangladesh & 0.76 & 0.00 & regression  \\
114 & socmob               & 0.91 & 0.00 & regression  \\
115 & space\_ga\_507       & 0.67 & 0.65 & regression  \\
116 & stock                & 0.99 & 0.99 & regression  \\
117 & sulfur               & 0.89 & 0.84 & regression  \\
118 & us\_crime            & 0.68 & 0.00 & regression  \\
119 & weather\_izmir       & 0.99 & 0.99 & regression  \\
120 & wind\_503            & 0.81 & 0.79 & regression  \\
121 & witmer\_census\_1980 & 0.50 & 0.00 & regression  \\
\bottomrule
\end{tabular}

\label{tab:detailed_scores_volcano}
\end{table*}

Tables \ref{tab:detailed_scores} and \ref{tab:detailed_scores_volcano} show the detailed macro F1 and $R^2$ scores for all systems on all benchmark datasets.

\end{document}